\documentclass{article}

\usepackage[final]{nips_2016}
\usepackage{float}

\usepackage[utf8]{inputenc} 
\usepackage[T1]{fontenc}    
\usepackage{hyperref}       
\usepackage{url}            
\usepackage{booktabs}       
\usepackage{amsfonts}       
\usepackage{nicefrac}       
\usepackage{microtype}      
\usepackage{graphicx} 
\usepackage{caption}
\usepackage[titletoc,title]{appendix}
\usepackage{subfigure}
\usepackage{amsmath,epsfig}
\usepackage[titletoc,title]{appendix}

\title{Adversarial Autoencoders}

\author{
Alireza Makhzani\\
University of Toronto\\
\texttt{makhzani@psi.toronto.edu}
\And
Jonathon Shlens \& Navdeep Jaitly \\
Google Brain \\
\texttt{\{shlens,ndjaitly\}@google.com} \\
\And
Ian Goodfellow \\
OpenAI \\
\texttt{goodfellow.ian@gmail.com } \\
\And
Brendan Frey\\
University of Toronto\\
\texttt{frey@psi.toronto.edu}
}

\begin{document}

\maketitle
\begin{abstract}
In this paper, we propose the ``adversarial autoencoder'' (AAE), which is a probabilistic autoencoder that uses the recently proposed generative adversarial networks (GAN) to perform variational inference by matching the aggregated posterior of the hidden code vector of the autoencoder with an arbitrary prior distribution. Matching the aggregated posterior to the prior ensures that generating from any part of prior space results in meaningful samples.  As a result, the decoder of the adversarial autoencoder learns a deep generative model that maps the imposed prior to the data distribution. We show how the adversarial autoencoder can be used in applications such as semi-supervised classification, disentangling style and content of images, unsupervised clustering, dimensionality reduction and data visualization. We performed experiments on MNIST, Street View House Numbers and Toronto Face datasets and show that adversarial autoencoders achieve competitive results in generative modeling and semi-supervised classification tasks.
\end{abstract}

\section{Introduction}
Building scalable generative models to capture rich distributions such as audio, images or video is one of the central challenges of machine learning.
Until recently, deep generative models, such as Restricted Boltzmann Machines (RBM), Deep Belief Networks (DBNs) and Deep Boltzmann Machines (DBMs) were
trained primarily by MCMC-based algorithms~\citep{geoff,russ}.
In these approaches the MCMC methods compute the gradient of log-likelihood which becomes more imprecise as training progresses. This is because
samples from the Markov Chains are unable to mix between modes fast enough.  In recent years, generative models have been developed that may be
trained via direct back-propagation and avoid the difficulties that come with MCMC training.
For example, variational autoencoders (VAE) \citep{vae,rezende} or importance weighted autoencoders \citep{yuri} use a recognition network to predict the posterior distribution over the latent variables, generative adversarial networks (GAN) \citep{gan} use an adversarial training procedure to directly shape the output distribution of the network via back-propagation and generative moment matching networks (GMMN) \citep{gmmn} use a moment matching cost function to learn the data distribution.

In this paper, we propose a general approach, called an adversarial autoencoder (AAE) that can turn an autoencoder into a generative model.
In our model, an autoencoder is trained with dual objectives -- a traditional reconstruction error criterion, and an adversarial training
criterion \citep{gan} that matches the aggregated posterior distribution of the latent representation of the autoencoder to an arbitrary prior distribution.
We show that this training criterion has a strong connection to VAE training.  The result of the training is that the encoder learns to
convert the data distribution to the prior distribution, while the decoder learns a deep generative model that maps the imposed prior to the data distribution.

\subsection{Generative Adversarial Networks}
The Generative Adversarial Networks (GAN) \citep{gan} framework establishes a min-max adversarial game between two neural networks -- a generative model, $G$,
and a discriminative model, $D$.  The discriminator model, $D(\mathbf{x})$, is a neural network that computes the probability that a point $\mathbf{x}$ in
data space is a sample from the data distribution (positive samples) that we are trying to model, rather than a sample from our generative model (negative samples).
Concurrently, the generator uses a function $G(\mathbf{z})$ that maps samples $\mathbf{z}$ from the prior $p(\mathbf{z})$ to the data space. $G(\mathbf{z})$ is
trained to maximally confuse the discriminator into believing that samples it generates come from the data distribution. The generator is trained by leveraging
the gradient of $D(\mathbf{x})$ w.r.t. $\mathbf{x}$, and using that to modify its parameters.
The solution to this game can be expressed as following~\citep{gan}:
$$\underset{G}{\min}   \text{ } \underset{D}{\max} \text{ } \text{E}_{\text{x} \sim p_{\text{data}}} [\text{log} D(\mathbf{x})] + \text{E}_{\mathbf{z} \sim p(\mathbf{z})} [\text{log} (1 - D(G(\mathbf{\mathbf{z}}))]$$

The generator $G$ and the discriminator $D$ can be found using alternating SGD in two stages: (a) Train the discriminator to distinguish the true samples from the fake samples generated by the generator.
(b) Train the generator so as to fool the discriminator with its generated samples.

\section{Adversarial Autoencoders}

\begin{figure}
\begin{center}
\hspace{4cm}\includegraphics[scale=.4]{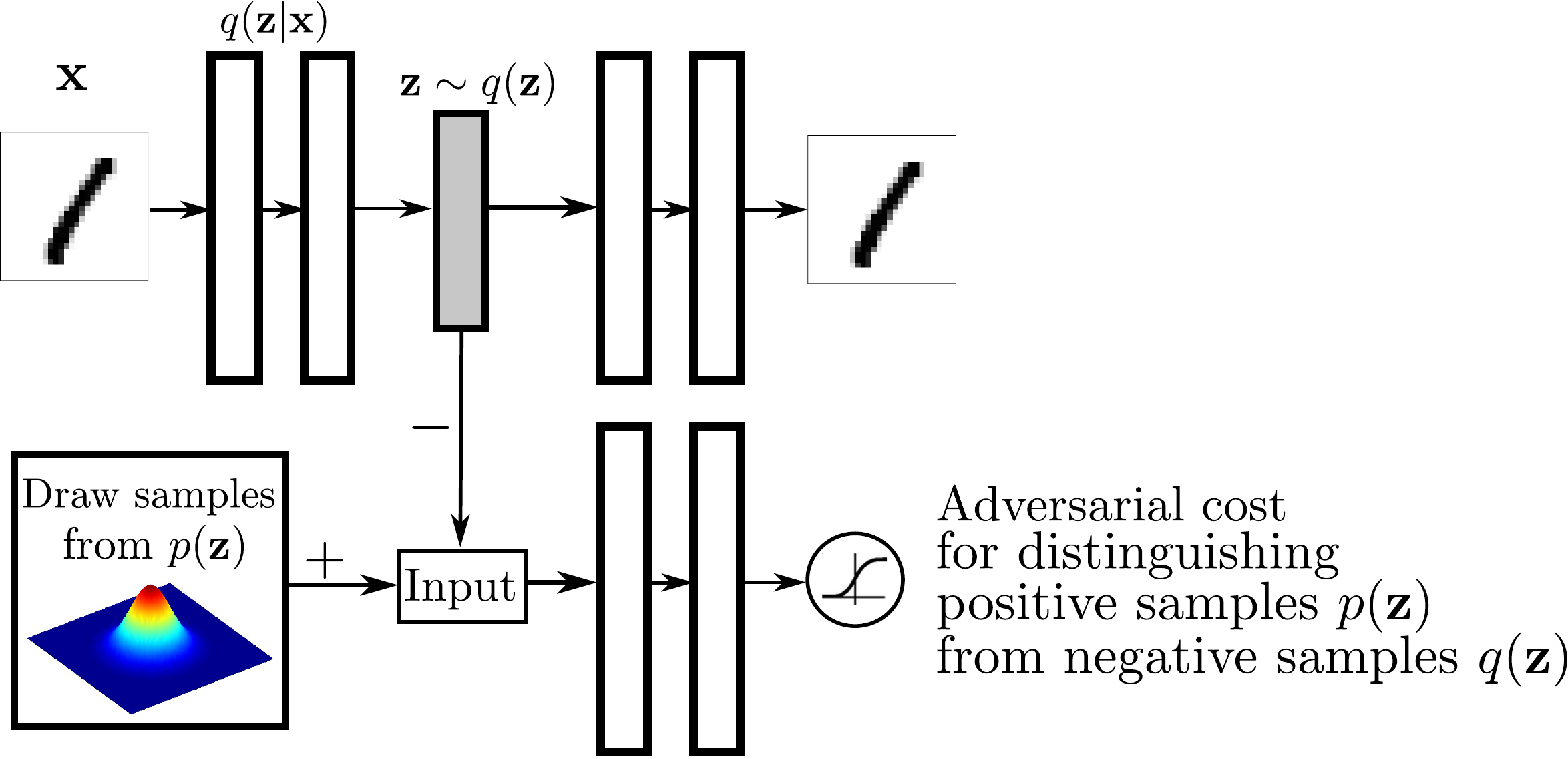}
\end{center}
\caption{\label{fig_adv_ae}Architecture of an adversarial autoencoder.
The top row is a standard autoencoder that reconstructs an image $\mathbf{x}$ from a latent code $\mathbf{z}$.
The bottom row diagrams a second network trained to discriminatively predict whether a sample arises from the hidden code of the autoencoder or from a sampled distribution specified by the user.}
\end{figure}

Let $\mathbf{x}$ be the input and $\mathbf{z}$ be the latent code vector (hidden units) of an autoencoder with a deep encoder and decoder.
Let $p(\mathbf{z})$ be the prior distribution we want to impose on the codes, $q( \mathbf{z} | \mathbf{x})$ be an encoding distribution
and $p( \mathbf{x} | \mathbf{z})$ be the decoding distribution.
Also let $p_{d} (\mathbf{x})$ be the data distribution, and $p(\mathbf{x})$ be the model distribution.
The encoding function of the autoencoder $q( \mathbf{z} | \mathbf{x})$ defines an aggregated posterior distribution of $q(\mathbf{z})$ on the hidden code vector of the autoencoder as follows:
\begin{equation}\label{qz} q(\mathbf{z}) = \int_\mathbf{x} q(\mathbf{z}|\mathbf{x}) p_{d} (\mathbf{x}) d\mathbf{x}
\end{equation}
The adversarial autoencoder is an autoencoder that is regularized by matching the aggregated posterior, $q(\mathbf{z})$, to an arbitrary prior, $p(\mathbf{z})$.
In order to do so, an adversarial network is attached on top of the hidden code vector of the autoencoder as illustrated in Figure \ref{fig_adv_ae}.
It is the adversarial network that guides $q(\mathbf{z})$ to match $p(\mathbf{z})$. The autoencoder, meanwhile, attempts to minimize the
reconstruction error.  The generator of the adversarial network is also the encoder of the autoencoder $q(\mathbf{z}|\mathbf{x})$.
The encoder ensures the aggregated posterior distribution can fool the discriminative adversarial network into thinking that the hidden code $q(\mathbf{z})$ comes from the true prior distribution $p(\mathbf{z})$.

Both, the adversarial network and the autoencoder are trained jointly with SGD in two phases -- the \emph{reconstruction} phase and the \emph{regularization}
phase -- executed on each mini-batch.  In the reconstruction phase, the autoencoder updates the encoder and the decoder to minimize the reconstruction error
of the inputs. In the regularization phase, the adversarial network first updates its discriminative network to tell apart the true samples (generated
using the prior) from the generated samples (the hidden codes computed by the autoencoder). The adversarial network then updates its generator (which is also the encoder of the autoencoder)
to confuse the discriminative network.

Once the training procedure is done, the decoder of the autoencoder will define a generative model that maps the imposed prior of $p(\mathbf{z})$ to the data distribution.

There are several possible choices for the encoder, $q(\mathbf{z}|\mathbf{x})$, of adversarial autoencoders:

{\bf Deterministic:} Here we assume that $q(\mathbf{z}|\mathbf{x})$ is a deterministic function of $\mathbf{x}$. In this case, the encoder is similar to the encoder of a standard autoencoder and the only source of stochasticity in $q(\mathbf{z})$ is the data distribution, $p_{d} (\mathbf{x})$.

{\bf Gaussian posterior:} Here we assume that $q(\mathbf{z}|\mathbf{x})$ is a Gaussian distribution whose mean and variance is predicted by the encoder network: $z_i \sim  \mathcal{N}(\mu_i(\mathbf{x}),\sigma_i(\mathbf{x}))$. In this case, the stochasticity in $q(\mathbf{z})$ comes from both the data-distribution and the randomness of the Gaussian distribution at the output of the encoder. We can use the same re-parametrization trick of \citep{vae} for back-propagation through the encoder network.

{\bf Universal approximator posterior:} Adversarial autoencoders can be used to train the $q(\mathbf{z}|\mathbf{x})$ as the universal approximator of the posterior. Suppose the encoder network of the adversarial autoencoder is the function $f(\mathbf{x},\eta)$ that takes the input $\mathbf{x}$ and a random noise $\eta$ with a fixed distribution (e.g., Gaussian). We can sample from arbitrary posterior distribution $q(\mathbf{z}|\mathbf{x})$, by evaluating $f(\mathbf{x},\eta)$ at different samples of $\eta$. In other words, we can assume $q(\mathbf{z}|\mathbf{x},\eta)=\delta(\mathbf{z} - f(\mathbf{x},\eta))$ and the posterior $q(\mathbf{z}|\mathbf{x})$ and the aggregated posterior $q(\mathbf{z})$ are defined as follows:

\vspace{-.2cm}

$$
q(\mathbf{z}|\mathbf{x}) = \int_\eta q(\mathbf{z}|\mathbf{x},\eta) p_{\eta}(\eta) d\eta \quad \Rightarrow  \quad q(\mathbf{z}) = \int_\mathbf{x} \int_\eta q(\mathbf{z}|\mathbf{x},\eta) p_d(\mathbf{x}) p_{\eta}(\eta) d\eta d\mathbf{x}
$$
\vspace{-.2cm}

In this case, the stochasticity in $q(\mathbf{z})$ comes from both the data-distribution and the random noise $\eta$ at the input of the encoder.
Note that in this case the posterior $q(\mathbf{z}|\mathbf{x})$ is no longer constrained to be Gaussian and the encoder can learn any arbitrary posterior distribution for a given input $\mathbf{x}$. Since there is an efficient method of sampling from the aggregated posterior $q(\mathbf{z})$, the adversarial training procedure can match $q(\mathbf{z})$ to $p(\mathbf{z})$ by direct back-propagation through the encoder network $f(\mathbf{x},\eta)$.

Choosing different types of $q(\mathbf{z}|\mathbf{x})$ will result in different kinds of models with different training dynamics. For example, in the deterministic case of $q(\mathbf{z}|\mathbf{x})$, the network has to match $q(\mathbf{z})$ to $p(\mathbf{z})$ by only exploiting the stochasticity of the data distribution, but since the empirical distribution of the data is fixed by the training set, and the mapping is deterministic, this might produce a $q(\mathbf{z})$ that is not very smooth. However, in the Gaussian or universal approximator case, the network has access to additional sources of stochasticity that could help it in the adversarial regularization stage by smoothing out $q(\mathbf{z})$. Nevertheless, after extensive hyper-parameter search, we obtained similar test-likelihood with each type of $q(\mathbf{z}|\mathbf{x})$. So in the rest of the paper, we only report results with the deterministic version of $q(\mathbf{z}|\mathbf{x})$.

\subsection{Relationship to Variational Autoencoders}

Our work is similar in spirit to variational autoencoders \citep{vae}; however, while they use a KL divergence penalty to impose a prior distribution on the hidden code vector of the autoencoder, we use an adversarial training procedure to do so by matching the aggregated posterior of the hidden code vector with the prior distribution.

VAE \citep{vae} minimizes the following upper-bound on the negative log-likelihood of $\mathbf{x}$:

\begin{equation} \label{vae}
\begin{split}
E_{\mathbf{x} \sim p_d(\mathbf{x})}[-\text{log } p(\mathbf{x})] &< E_{\mathbf{x}}[\text{E}_{q(\mathbf{z}|\mathbf{x})} [-\text{log}(p(\text{x}|\text{z})]] + E_{\mathbf{x}} [\text{KL}( q(\mathbf{z}|\mathbf{x}) \| p(\mathbf{z}) )]\\
  &=  E_{\mathbf{x}}[\text{E}_{q(\mathbf{z}|\mathbf{x})} [-\text{log }p(\text{x}|\text{z})]] - E_{\mathbf{x}}[H({q(\mathbf{z}|\mathbf{x})})] + E_{q(\mathbf{z})}[-\text{log } p(\mathbf{z})] \\
  &=  E_{\mathbf{x}}[\text{E}_{q(\mathbf{z}|\mathbf{x})} [-\text{log }p(\text{x}|\text{z})]] - E_{\mathbf{x}}[\sum_i \text{log} \sigma_i(\mathbf{x}))] + E_{q(\mathbf{z})}[-\text{log } p(\mathbf{z})] + \text{const.}\\
 &= \text{Reconstruction} - \text{Entropy} + \text{CrossEntropy}(q(\mathbf{z}),p(\mathbf{z}))
\end{split}
\end{equation}

where the aggregated posterior $q(\mathbf{z})$ is defined in Eq. (\ref{qz}) and we have assumed $q(\mathbf{z}|\mathbf{x})$ is Gaussian and $p(\mathbf{z})$ is an arbitrary distribution.
The variational bound contains three terms.
The first term can be viewed as the reconstruction term of an autoencoder and the second and third terms can be viewed as regularization terms.
Without the regularization terms, the model is simply a standard autoencoder that reconstructs the input.
However, in the presence of the regularization terms, the VAE learns a latent representation that is compatible with $p(\mathbf{z})$.
The second term of the cost function encourages large variances for the posterior distribution while the third term minimizes the cross-entropy between the aggregated posterior $q(\mathbf{z})$ and the prior $p(\mathbf{z})$. 
KL divergence or the cross-entropy term in Eq. (\ref{vae}), encourages $q(\mathbf{z})$ to pick the modes of $p(\mathbf{z})$. In adversarial autoencoders, we replace the second two terms with an adversarial training procedure that encourages $q(\mathbf{z})$ to match to the whole distribution of $p(\mathbf{z})$.

\begin{figure}[t]
\centering\includegraphics[scale=.13]{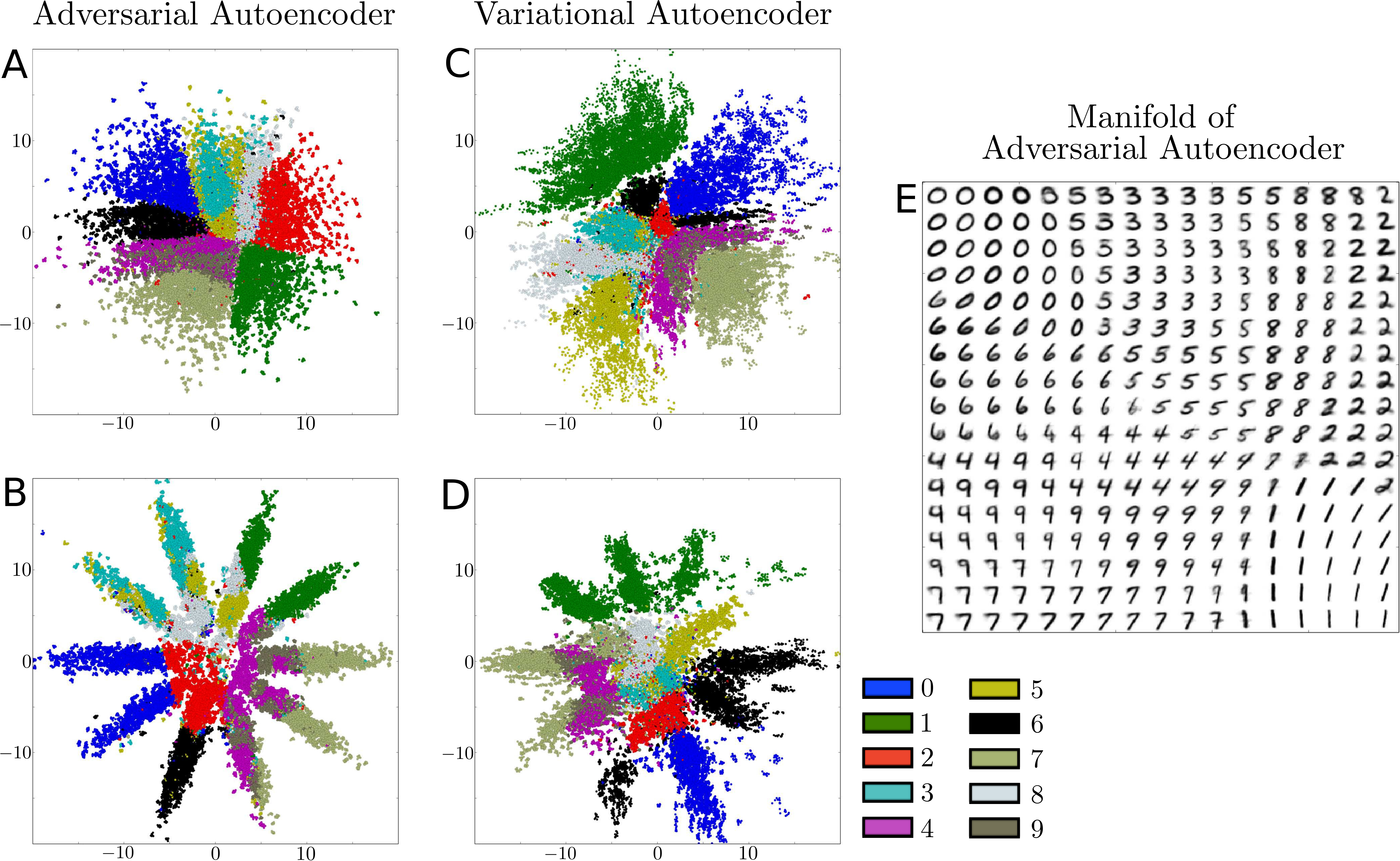}
\caption{\label{fig_mnist}Comparison of adversarial and variational autoencoder on MNIST. The hidden code $\mathbf{z}$ of the \emph{hold-out} images for an adversarial autoencoder fit to (a) a 2-D Gaussian and (b) a mixture of 10 2-D Gaussians. Each color represents the associated label. Same for variational autoencoder with (c) a 2-D gaussian and (d) a mixture of 10 2-D Gaussians. (e) Images generated by uniformly sampling the Gaussian percentiles along each hidden code dimension $\mathbf{z}$ in the 2-D Gaussian adversarial autoencoder.}
\end{figure}

In this section, we compare the ability of the adversarial autoencoder to the VAE to impose a specified prior distribution $p(\mathbf{z})$ on the coding distribution.
Figure \ref{fig_mnist}a shows the coding space $\mathbf{z}$ of the test data resulting from an adversarial autoencoder trained on MNIST digits in which a
spherical 2-D Gaussian prior distribution is imposed on the hidden codes $\mathbf{z}$.
The learned manifold in Figure \ref{fig_mnist}a exhibits sharp transitions indicating that the coding space is filled and exhibits no ``holes''.
In practice, sharp transitions in the coding space indicate that images generated by interpolating within $\mathbf{z}$ lie on the data manifold (Figure \ref{fig_mnist}e).
By contrast, Figure \ref{fig_mnist}c shows the coding space of a VAE with the same architecture used in the adversarial autoencoder experiments.
We can see that in this case the VAE roughly matches the shape of a 2-D Gaussian distribution.
However, no data points map to several local regions of the coding space indicating that the VAE may not have captured the data manifold
as well as the adversarial autoencoder.

Figures \ref{fig_mnist}b and \ref{fig_mnist}d show the code space of an adversarial autoencoder and of a VAE where the imposed distribution is a mixture of 10 2-D Gaussians.
The adversarial autoencoder successfully matched the aggregated posterior with the prior distribution (Figure \ref{fig_mnist}b).
In contrast, the VAE exhibit systematic differences from the mixture 10 Gaussians indicating that the VAE emphasizes matching the modes of the distribution as discussed above (Figure \ref{fig_mnist}d).

An important difference between VAEs and adversarial autoencoders is that in VAEs, in order to back-propagate through the KL divergence by Monte-Carlo sampling, we need to have access to the exact functional form of the prior distribution.
However, in AAEs, we only need to be able to sample from the prior distribution in order to induce $q(\mathbf{z})$ to match $p(\mathbf{z})$. In Section \ref{semi2}, we will demonstrate that the adversarial autoencoder can impose complicated distributions (e.g., swiss roll distribution) without having access to the explicit functional form of the distribution.

\subsection{Relationship to GANs and GMMNs}\label{gmmn}

In the original generative adversarial networks (GAN) paper~\citep{gan}, GANs were used to impose the data distribution at the pixel level on the output layer of a neural network. Adversarial autoencoders, however, rely on the autoencoder training to capture the data distribution. In adversarial training procedure of our method, a much simpler distribution (e.g., Gaussian as opposed to the data distribution) is imposed in a much lower dimensional space (e.g., $20$ as opposed to $1000$) which results in a better test-likelihood as is discussed in Section~\ref{experiments}.

Generative moment matching networks (GMMN) \citep{gmmn} use the maximum mean discrepancy (MMD) objective to shape the distribution of the output layer of a neural network. The MMD objective can be interpreted as minimizing the distance between all moments of the model distribution and the data distribution. It has been shown that GMMNs can be combined with pre-trained dropout autoencoders to achieve better likelihood results (GMMN+AE). Our adversarial autoencoder also relies on the autoencoder to capture the data distribution. However, the main difference of our work with GMMN+AE is that the adversarial training procedure of our method acts as a regularizer that shapes the code distribution while training the autoencoder from scratch; whereas, the GMMN+AE model first trains a standard dropout autoencoder and then fits a distribution in the code space of the pre-trained network. In Section \ref{experiments}, we will show that the test-likelihood achieved by the joint training scheme of adversarial autoencoders outperforms the test-likelihood of GMMN and GMMN+AE on MNIST and Toronto Face datasets.

\subsection{Incorporating Label Information in the Adversarial Regularization}\label{semi2}
In the scenarios where data is labeled, we can incorporate the label information in the adversarial training stage to better shape the distribution of the hidden code. 
In this section, we describe how to leverage partial or complete label information to regularize the latent representation of the autoencoder more heavily.
To demonstrate this architecture we return to Figure \ref{fig_mnist}b in which the adversarial autoencoder is fit to a mixture of 10 2-D Gaussians.
We now aim to force each mode of the mixture of Gaussian distribution to represent a single label of MNIST.

\begin{figure}[b]
\begin{center}
\centering 
\hspace{0.2cm}\includegraphics[scale=.35]{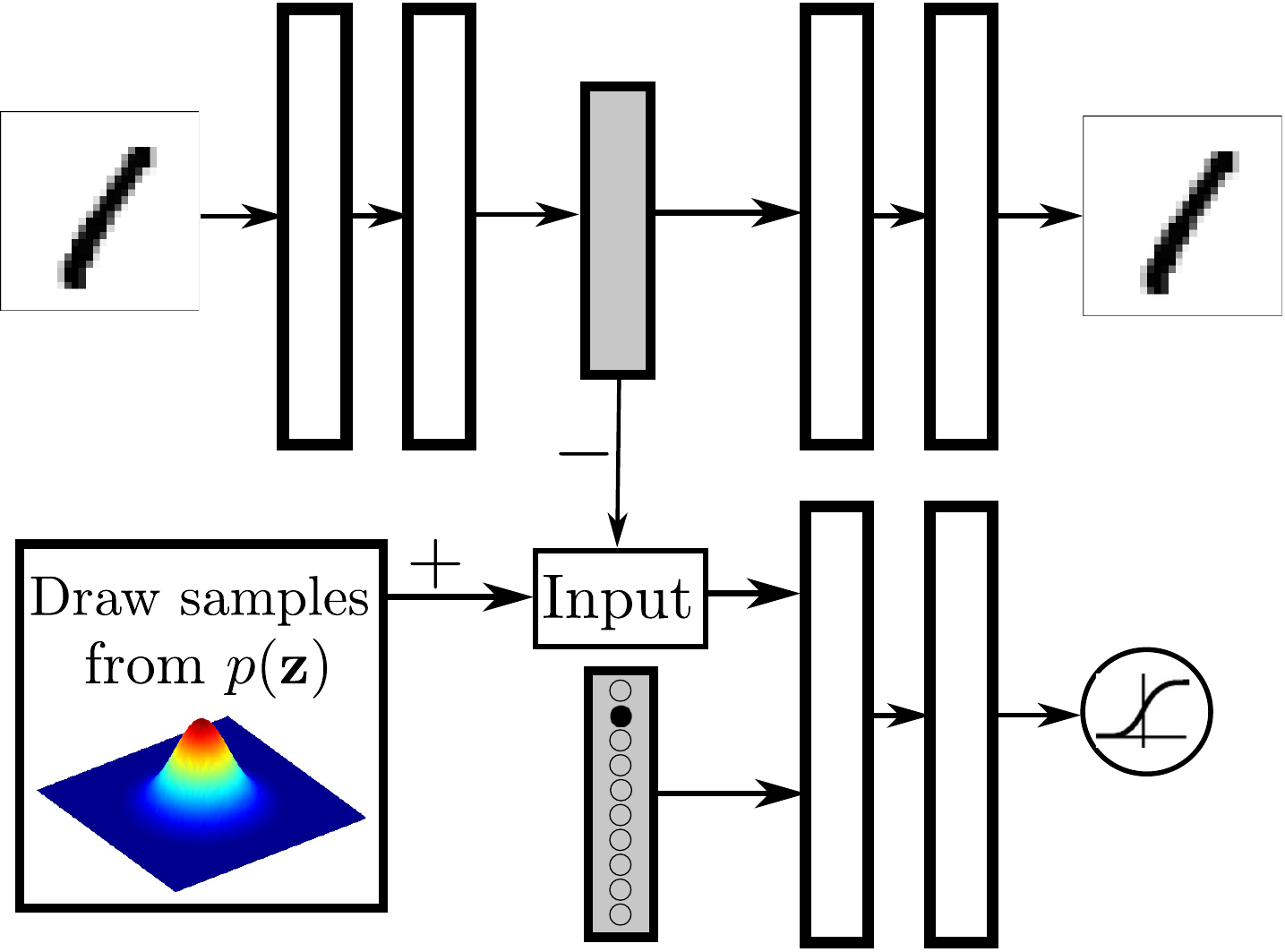}
\caption{\label{fig_semi_adv_b}Regularizing the hidden code by providing a one-hot vector to the discriminative network. The one-hot vector has an extra label for training points with unknown classes.}
\end{center}
\end{figure}

\begin{figure}[t]
\centering\includegraphics[scale=.1]{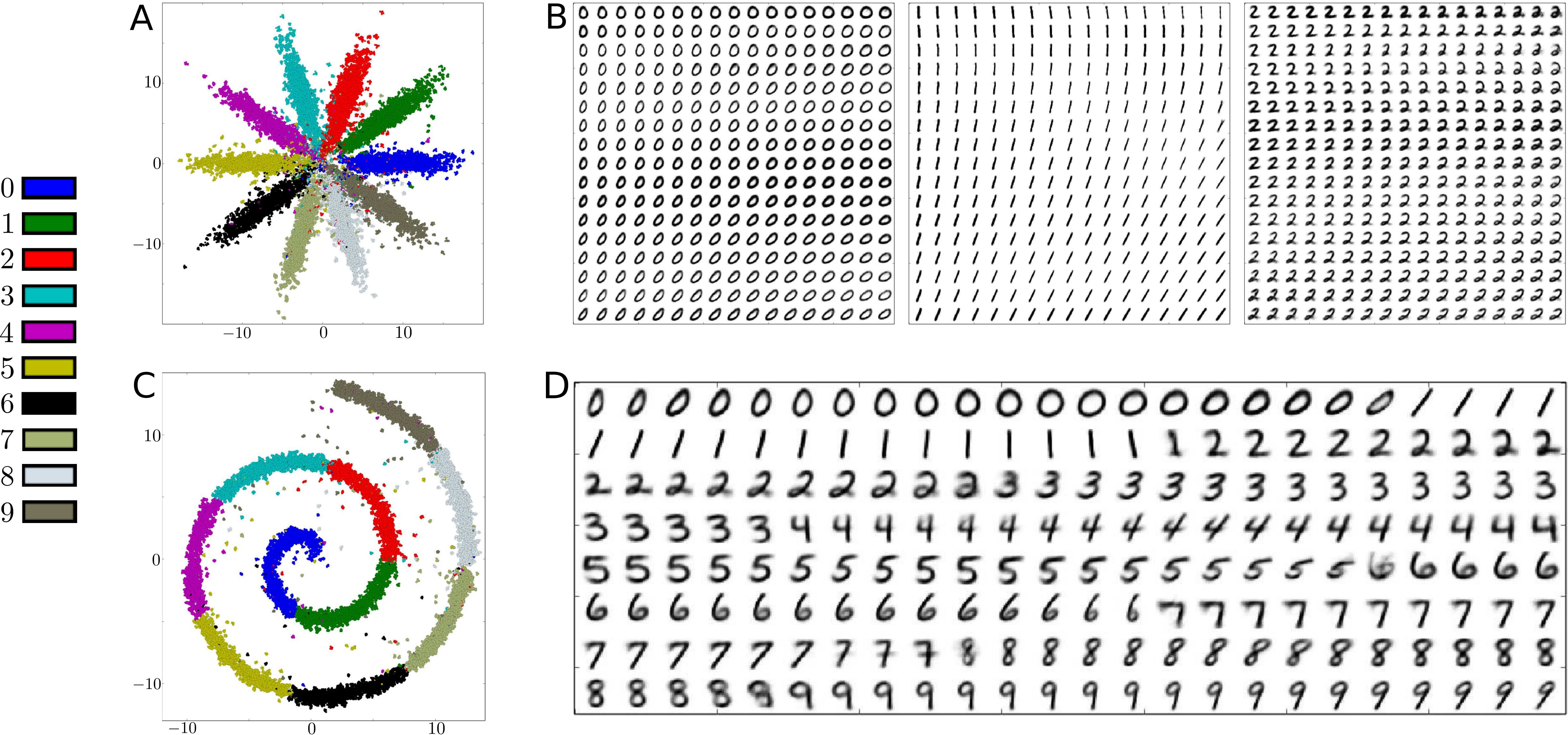}
\caption{\label{fig_semi_regularization}Leveraging label information to better regularize the hidden code. \textbf{Top Row:} Training the coding space to match a mixture of 10 2-D Gaussians: (a) Coding space $\mathbf{z}$ of the \emph{hold-out} images. (b) The manifold of the first 3 mixture components: each panel includes images generated by uniformly sampling the Gaussian percentiles along the axes of the corresponding mixture component. \textbf{Bottom Row:} Same but for a swiss roll distribution (see text). Note that labels are mapped in a numeric order (i.e., the first 10\% of swiss roll is assigned to digit $0$ and so on): (c) Coding space $\mathbf{z}$ of the \emph{hold-out} images. (d) Samples generated by walking along the main swiss roll axis.}
\end{figure}

Figure \ref{fig_semi_adv_b} demonstrates the training procedure for this semi-supervised approach.
We add a one-hot vector to the input of the discriminative network to associate the label with a mode of the distribution. The one-hot vector acts as switch that selects the corresponding decision boundary of the discriminative network given the class label. This one-hot vector has an extra class for unlabeled examples.
For example, in the case of imposing a mixture of 10 2-D Gaussians (Figure \ref{fig_mnist}b and \ref{fig_semi_regularization}a), the one hot vector contains 11 classes. Each of the first 10 class selects a decision boundary for the corresponding individual mixture component. The extra class in the one-hot vector corresponds to unlabeled training points. When an unlabeled point is presented to the model, the extra class is turned on, to select the decision boundary for the full mixture of Gaussian distribution.
During the positive phase of adversarial training, we provide the label of the mixture component (that the positive sample is drawn from) to the discriminator through the one-hot vector.  The positive samples fed for unlabeled examples come from the full mixture of Gaussian, rather than from a particular class.
During the negative phase, we provide the label of the training point image to the discriminator through the one-hot vector.

Figure \ref{fig_semi_regularization}a shows the latent representation of an adversarial autoencoder trained with a prior that is a mixture of 10 2-D Gaussians trained on 10K labeled MNIST examples and 40K unlabeled MNIST examples.
In this case, the $i$-th mixture component of the prior has been assigned to the $i$-th class in a semi-supervised fashion.
Figure \ref{fig_semi_regularization}b shows the manifold of the first three mixture components.
Note that the style representation is consistently represented within each mixture component, independent of its class.
For example, the upper-left region of all panels in Figure \ref{fig_semi_regularization}b correspond to the upright writing style and lower-right region of these panels correspond to the tilted writing style of digits.

This method may be extended to arbitrary distributions with no parametric forms -- as demonstrated by mapping the MNIST data set onto a ``swiss roll'' (a conditional Gaussian distribution whose mean is uniformly distributed along the length of a swiss roll axis).
Figure \ref{fig_semi_regularization}c depicts the coding space $\mathbf{z}$ and Figure \ref{fig_semi_regularization}d highlights the images generated by walking along the swiss roll axis in the latent space.

\vspace{.2cm}
\section{Likelihood Analysis of Adversarial Autoencoders}\label{experiments}
\vspace{-.3cm}

\begin{center}
\begin{table}[t]
\small
\centering 
\begin{tabular}{ l  || c || c || c || c}
  \hline
  & MNIST (10K) & MNIST (10M) & TFD (10K) & TFD (10M)\\
  \hline

  DBN \citep{geoff} & $138 \pm 2$ & - & $1909 \pm 66$ & -\\
  Stacked CAE \citep{stacked_cae} & $121 \pm 1.6$ & - & $2110 \pm 50$ & -  \\
  Deep GSN \citep{gsn} & $214 \pm 1.1$ & - & $1890   \pm 29$ & - \\
  GAN \citep{gan} & $225 \pm 2$ & $386$ & $2057 \pm 26$ & -\\
  GMMN + AE \citep{gmmn} & $282 \pm 2$ & - & $2204 \pm 20$ & -\\
  Adversarial Autoencoder & $\mathbf{340} \pm \mathbf{2}$ & $\mathbf{427}$ & $\mathbf{2252} \pm \mathbf{16}$ & $\mathbf{2522}$\\
  \hline
\end{tabular}
\caption{\label{table:table_mnist}Log-likelihood of test data on MNIST and Toronto Face dataset. Higher values are better. On both datasets we report the Parzen window estimate of the log-likelihood obtained by drawing 10K or 10M samples from the trained model. For MNIST, we compare against other models on the real-valued version of the dataset.}
\end{table}
\end{center}

The experiments presented in the previous sections have only demonstrated qualitative results.
In this section we measure the ability of the AAE as a generative model to capture the data distribution by comparing the likelihood of this model to generate hold-out images on the MNIST and Toronto face dataset (TFD) using the evaluation procedure described in~\citep{gan}.

\begin{figure}[b]
\centering
\subfigure[MNIST samples (8-D Gaussian)]{
\includegraphics[scale=.22]{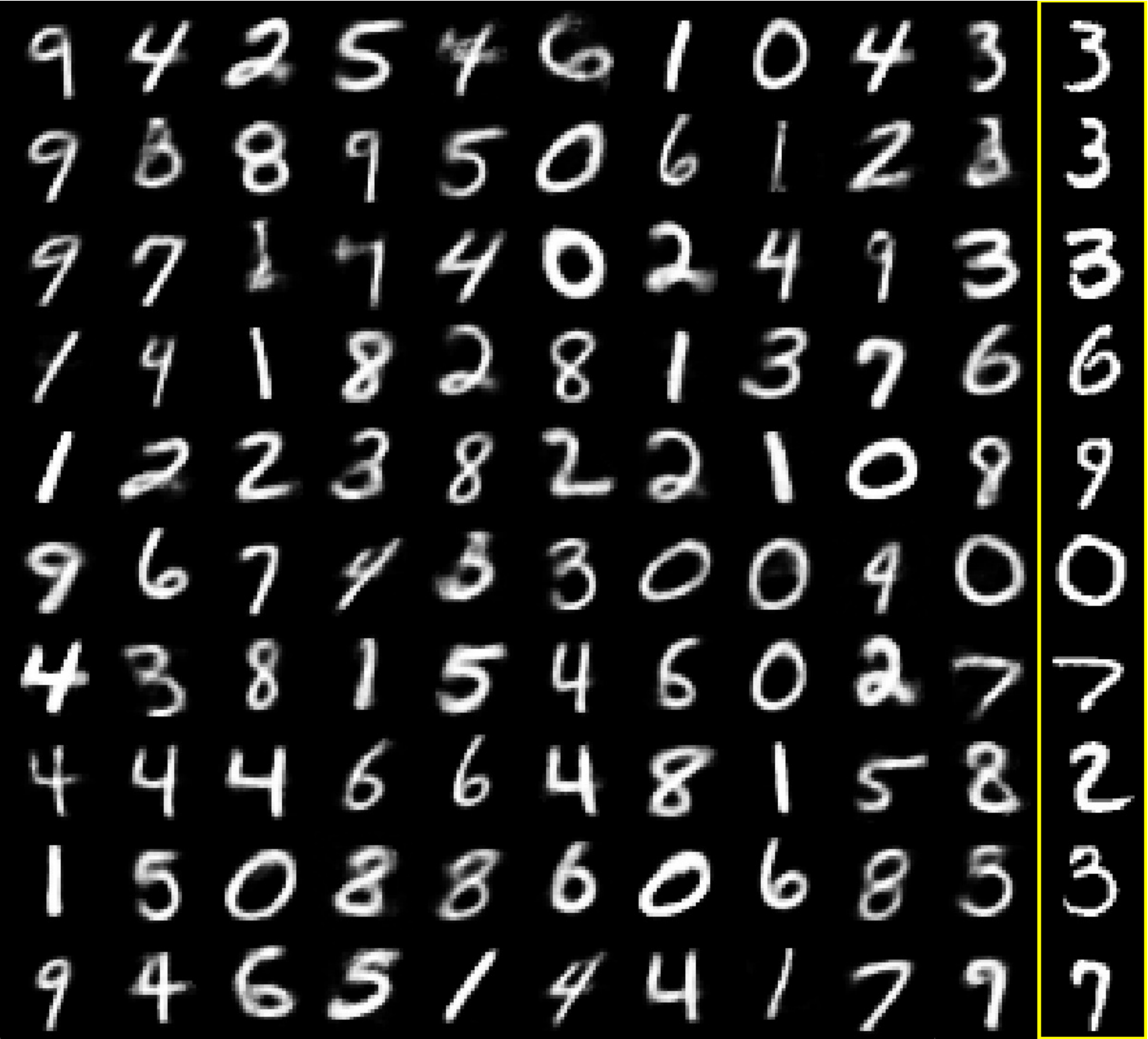}}
\hspace{.2cm}
\subfigure[TFD samples (15-D Gaussian)]{
\includegraphics[scale=.22]{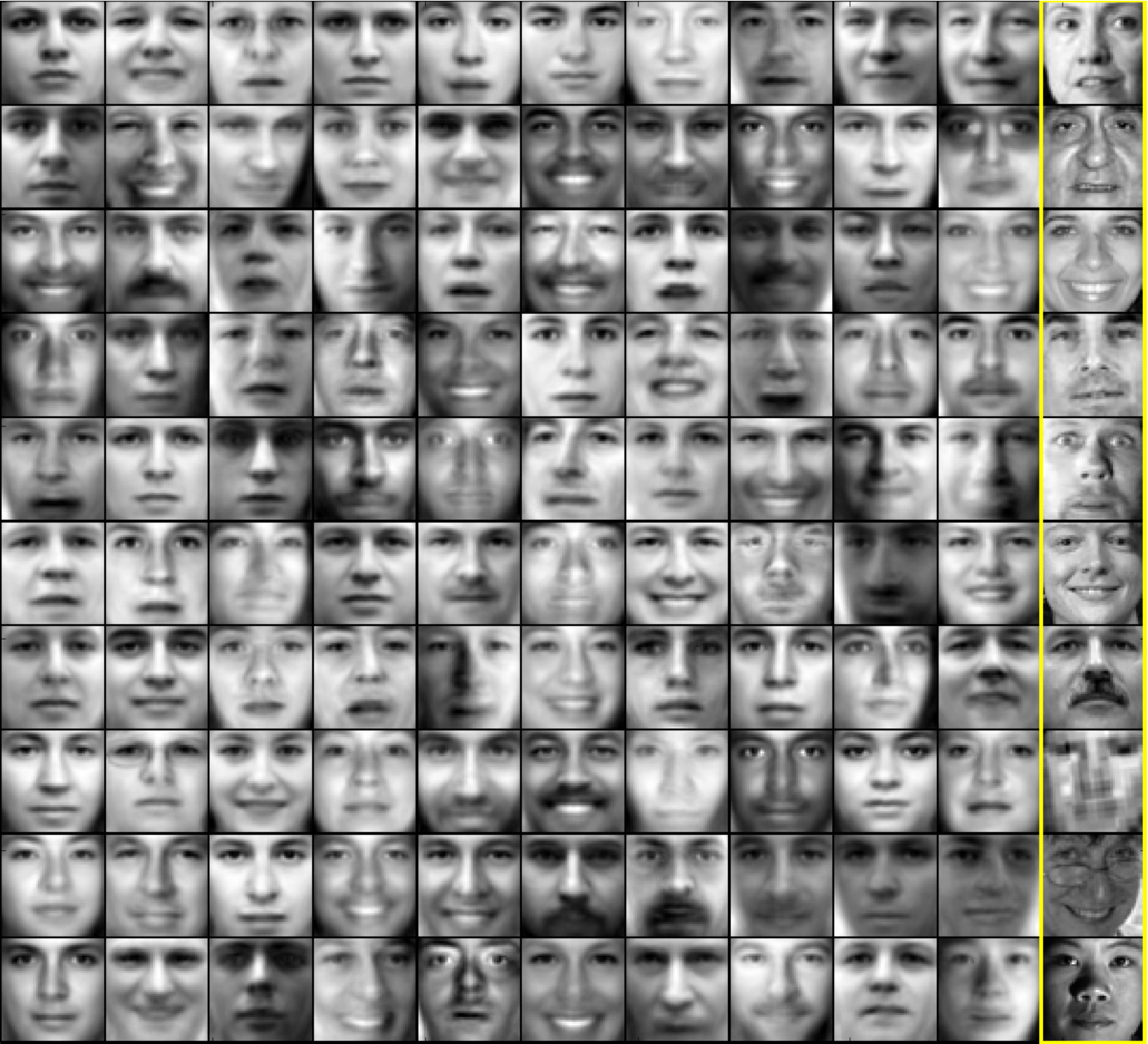}}
\caption{\label{fig_samples}Samples generated from an adversarial autoencoder trained on MNIST and Toronto Face dataset (TFD). The last column shows the closest training images in pixel-wise Euclidean distance to those in the second-to-last column.}
\end{figure}

We trained an adversarial autoencoder on MNIST and TFD in which the model imposed a high-dimensional Gaussian distribution on the underlying hidden code.
Figure \ref{fig_samples} shows samples drawn from the adversarial autoencoder trained on these datasets. 
A video showing the learnt TFD manifold can be found at \url{http://www.comm.utoronto.ca/~makhzani/adv_ae/tfd.gif}. 
To determine whether the model is over-fitting by copying the training data points, we used the last column of these figures to show the nearest neighbors, in Euclidean distance, to the generative model samples in the second-to-last column.

We evaluate the performance of the adversarial autoencoder by computing its log-likelihood on the hold out test set. Evaluation of the model using likelihood is not straightforward because we can not directly compute the probability of an image.
Thus, we calculate a lower bound of the true log-likelihood using the methods described in prior work~\citep{stacked_cae,gsn,gan}.
We fit a Gaussian Parzen window (kernel density estimator) to $10,000$ samples generated from the model and compute the likelihood of the test data under this distribution.
The free-parameter $\sigma$ of the Parzen window is selected via cross-validation.

Table \ref{table:table_mnist} compares the log-likelihood of the adversarial autoencoder for real-valued MNIST and TFD to many state-of-the-art methods including DBN \citep{geoff}, Stacked CAE \citep{stacked_cae}, Deep GSN \citep{gsn}, Generative Adversarial Networks \citep{gan} and GMMN + AE \citep{gmmn}.

Note that the Parzen window estimate is a lower bound on the true log-likelihood and the tightness of this bound depends on the number of samples drawn.
To obtain a comparison with a tighter lower bound, we additionally report Parzen window estimates evaluated with 10 million samples for both the adversarial autoencoders and the generative adversarial network~\citep{gan}.
In all comparisons we find that the adversarial autoencoder achieves superior log-likelihoods to competing methods.
However, the reader must be aware that the metrics currently available for evaluating the likelihood of generative models such as GANs are deeply flawed. 
Theis et al. \citep{theis} detail the problems with such metrics, including the 10K and 10M sample Parzen window estimate.

\section{Supervised Adversarial Autoencoders}\label{sec:supervised}
\begin{figure}[t]
\begin{center}
\centering 
\hspace{0.2cm}\includegraphics[scale=.35]{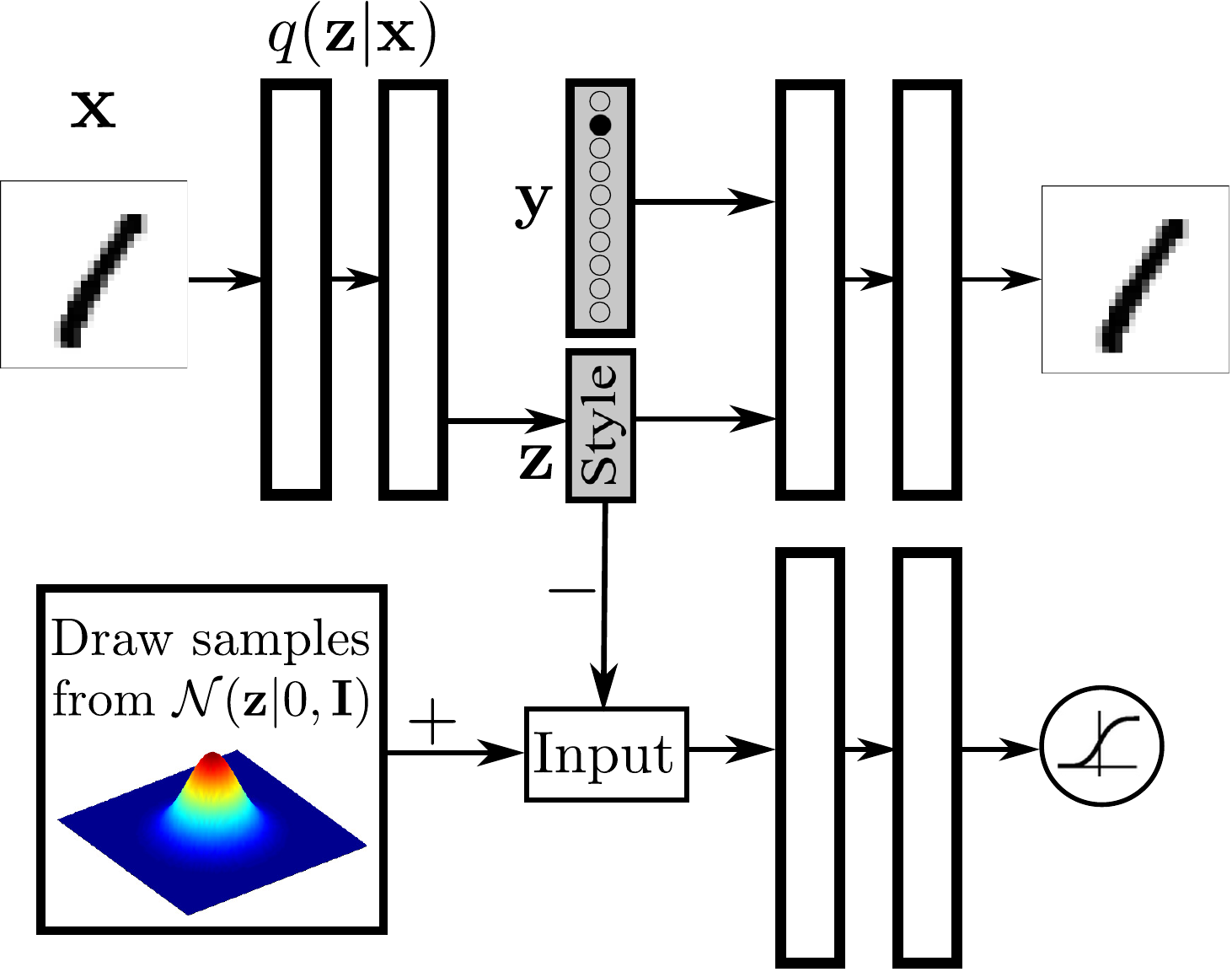}
\caption{\label{fig_semi_adv}Disentangling the label information from the hidden code by providing the one-hot vector to the generative model. The hidden code in this case learns to represent the style of the image.}
\end{center}
\end{figure}
Semi-supervised learning is a long-standing conceptual problem in machine learning. Recently, generative models have become one of the most popular approaches for semi-supervised learning as they can disentangle the class label information from many other latent factors of variation in a principled way \citep{semi-vae,adgm}.

In this section, we first focus on the fully supervised scenarios and discuss an architecture of adversarial autoencoders that can separate the class label information from the image style information. We then extend this architecture to the semi-supervised settings in Section \ref{sec:semi-supervised}.

\begin{figure}[b]
\centering
\subfigure[MNIST]{
\includegraphics[scale=.245]{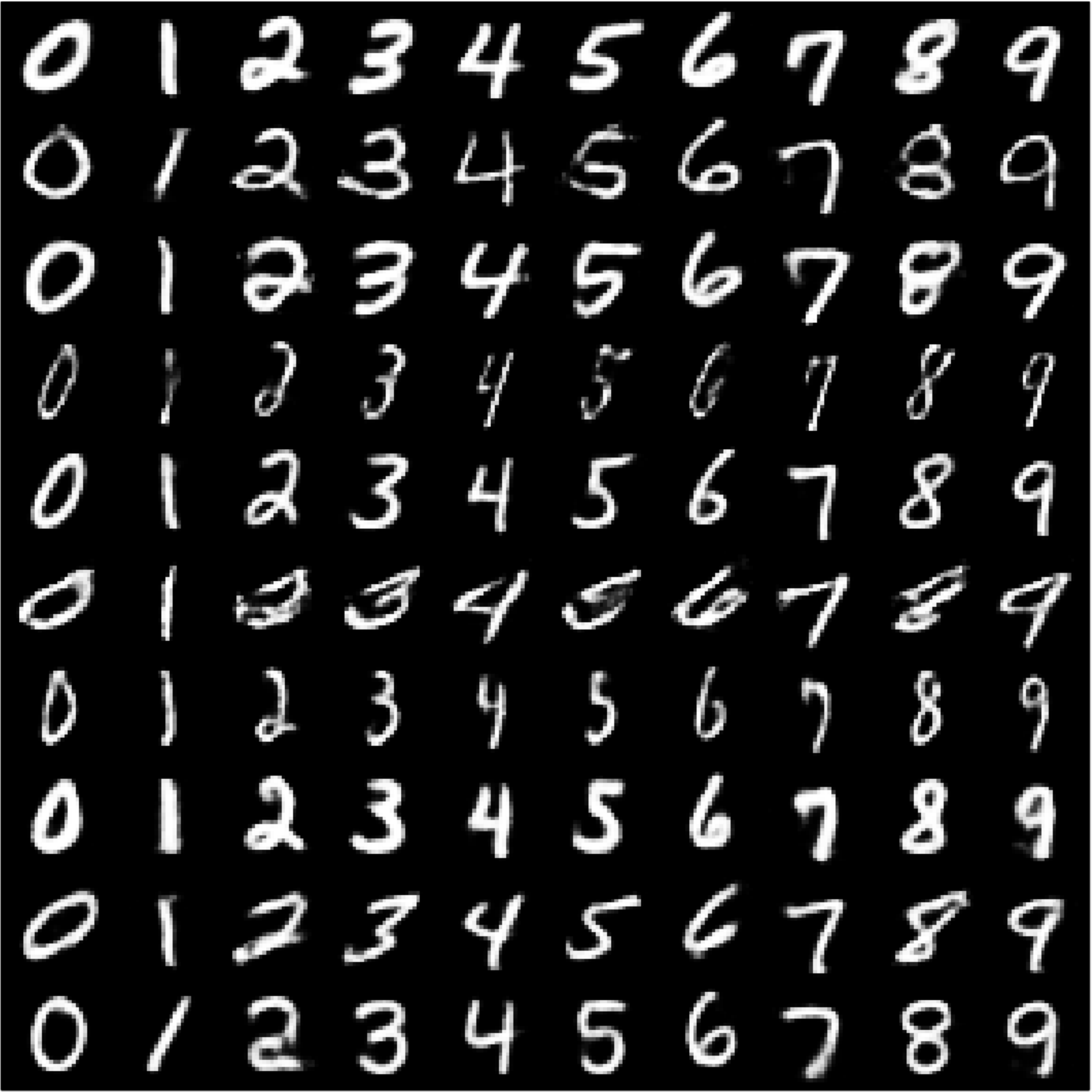}}
\subfigure[SVHN]{
\hspace{.2cm}\includegraphics[scale=.24]{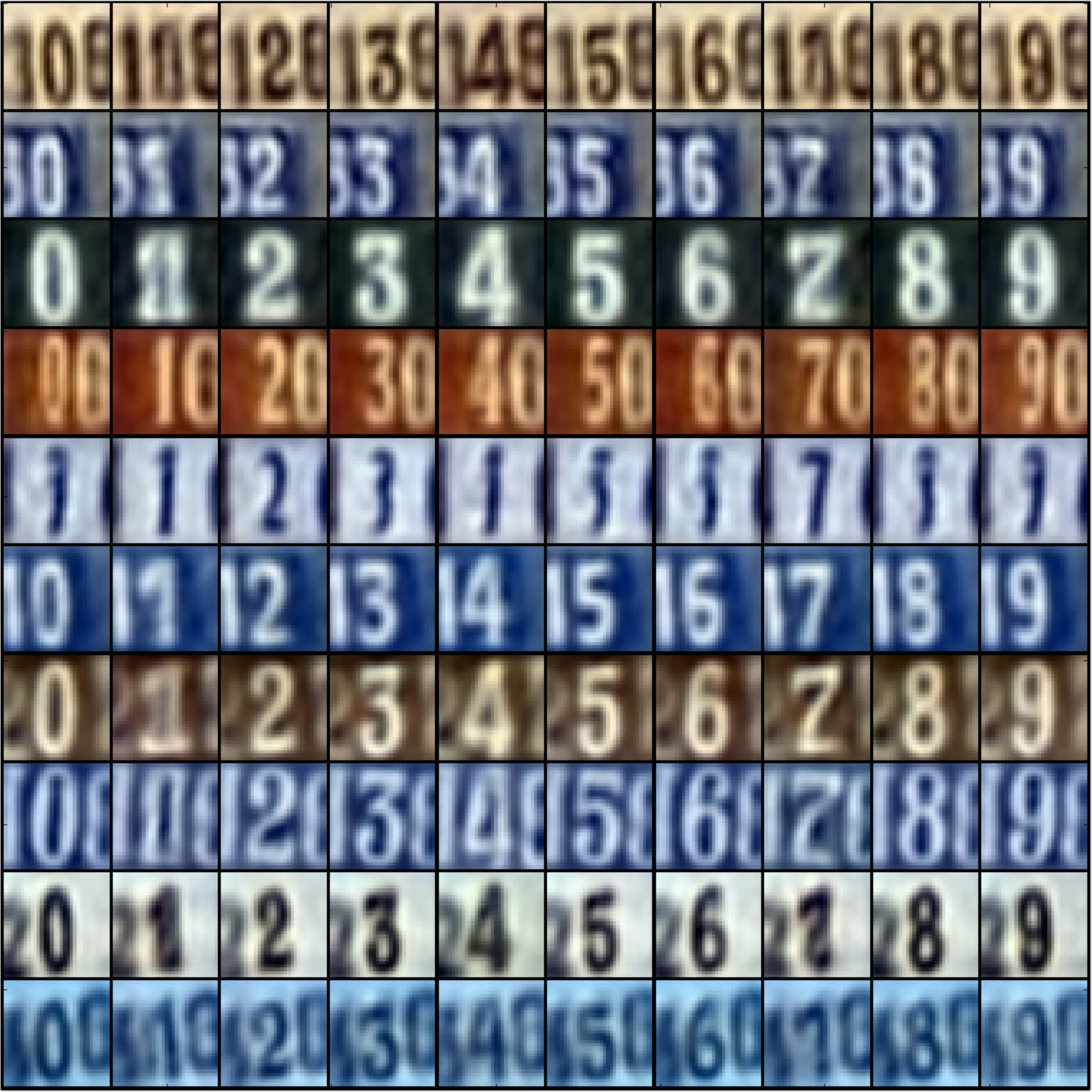}}
\caption{\label{fig_disentangle}Disentangling content and style (15-D Gaussian) on MNIST and SVHN datasets.}
\end{figure}

In order to incorporate the label information, we alter the network architecture of Figure \ref{fig_adv_ae} to provide a one-hot vector encoding of the label to the decoder (Figure \ref{fig_semi_adv}).
The decoder utilizes both the one-hot vector identifying the label and the hidden code $\mathbf{z}$ to reconstruct the image.
This architecture forces the network to retain all information independent of the label in the hidden code $\mathbf{z}$.

Figure \ref{fig_disentangle}a demonstrates the results of such a network trained on MNIST digits in which the hidden code is forced into a 15-D Gaussian.
Each row of Figure \ref{fig_disentangle}a presents reconstructed images in which the hidden code $\mathbf{z}$ is fixed to a particular value but the label is systematically explored.
Note that the style of the reconstructed images is consistent across a given row.
Figure \ref{fig_disentangle}b demonstrates the same experiment applied to Street View House Numbers dataset \citep{svhn}. A video showing the learnt SVHN style manifold can be found at \url{http://www.comm.utoronto.ca/~makhzani/adv_ae/svhn.gif}.
In this experiment, the one-hot vector represents the label associated with the \textit{central} digit in the image.
Note that the style information in each row contains information about the labels of the left-most and right-most digits because the left-most and right-most digits are not provided as label information in the one-hot encoding.

\section{Semi-Supervised Adversarial Autoencoders}\label{sec:semi-supervised}

\begin{figure}[b]
\begin{center}
\centering 
\includegraphics[scale=.4]{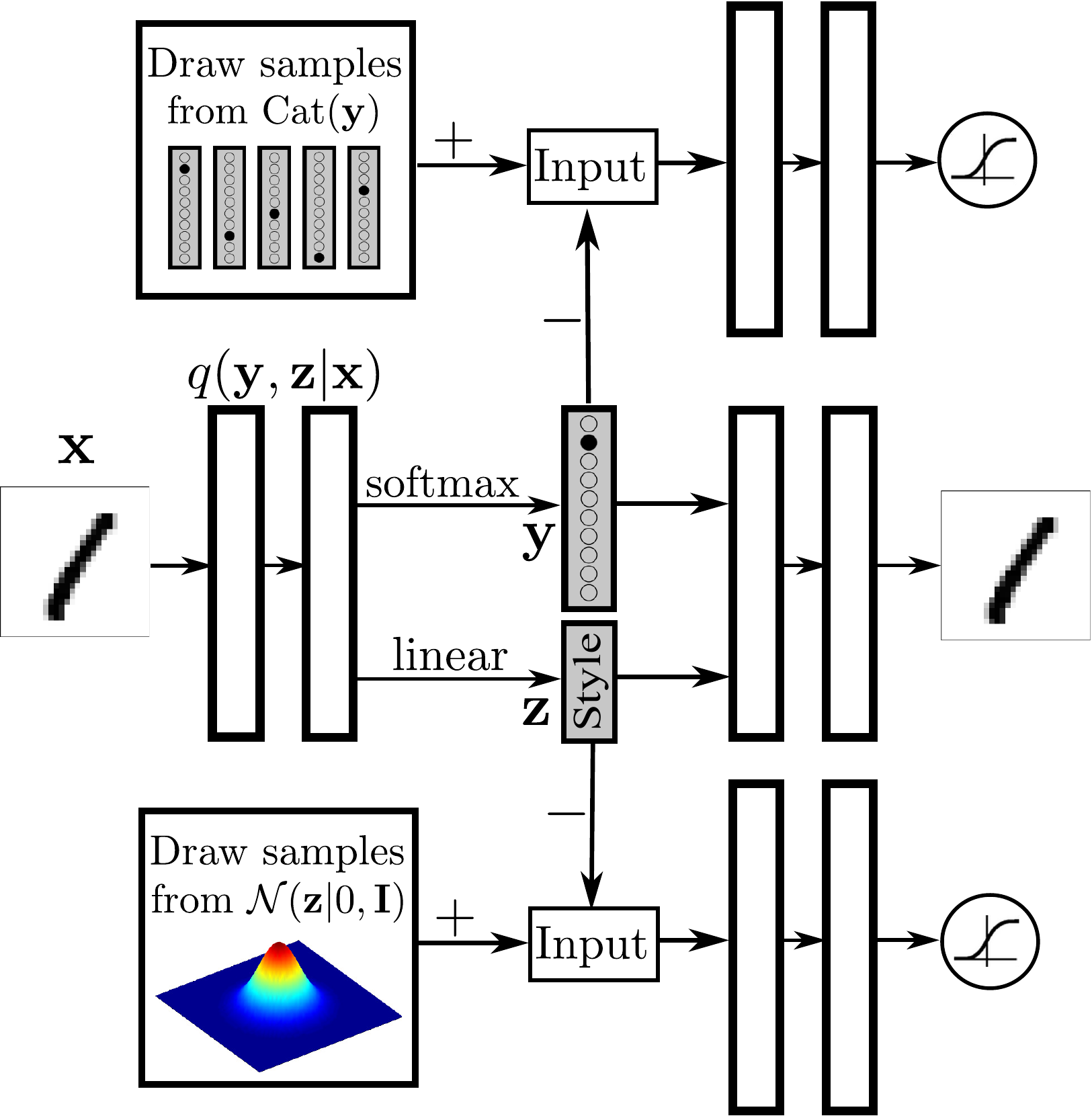}
\caption{\label{fig_semi_aae}Semi-Supervised AAE: the top adversarial network imposes a Categorical distribution on the label representation and the bottom adversarial network imposes a Gaussian distribution on the style representation. $q(\mathbf{y}|\mathbf{x})$ is trained on the labeled data in the semi-supervised settings. }
\end{center}
\end{figure}

Building on the foundations from Section \ref{sec:supervised}, we now use the adversarial autoencoder to develop models for semi-supervised learning that exploit the generative description of the unlabeled data to improve the classification performance that would be obtained by using only the labeled data. Specifically, we assume the data is generated by a latent class variable $\mathbf{y}$ that comes from a Categorical distribution as well as a continuous latent variable $\mathbf{z}$ that comes from a Gaussian distribution:

$$p(\mathbf{y}) = \text{Cat}(\mathbf{y}) \qquad  p(\mathbf{\mathbf{z}}) = \mathcal{N}(\mathbf{z}|0,\mathbf{I})$$

We alter the network architecture of Figure \ref{fig_semi_adv} so that the inference network of the AAE predicts both the discrete class variable $\mathbf{y}$ and the continuous latent variable $\mathbf{z}$ using the encoder $q(\mathbf{z},\mathbf{y}|\mathbf{x})$ (Figure \ref{fig_semi_aae}). The decoder then utilizes both the class label as a one-hot vector and the continuous hidden code $\mathbf{z}$ to reconstruct the image. There are two separate adversarial networks that regularize the hidden representation of the autoencoder. The first adversarial network imposes a Categorical distribution on the label representation. This adversarial network ensures that the latent class variable $\mathbf{y}$ does not carry any style information and that the aggregated posterior distribution of $\mathbf{y}$ matches the Categorical distribution. The second adversarial network imposes a Gaussian distribution on the style representation which ensures the latent variable $\mathbf{z}$ is a continuous Gaussian variable.

Both of the adversarial networks as well as the autoencoder are trained jointly with SGD in three phases -- the \emph{reconstruction} phase, \emph{regularization} phase and the \emph{semi-supervised classification} phase. In the reconstruction phase, the autoencoder updates the encoder $q(\mathbf{z},\mathbf{y}|\mathbf{x})$ and the decoder to minimize the reconstruction error of the inputs on an unlabeled mini-batch. In the regularization phase, each of the adversarial networks first updates their discriminative network to tell apart the true samples (generated using the Categorical and Gaussian priors) from the generated samples (the hidden codes computed by the autoencoder). The adversarial networks then update their generator to confuse their discriminative networks. In the semi-supervised classification phase, the autoencoder updates $q(\mathbf{y}|\mathbf{x})$ to minimize the cross-entropy cost on a labeled mini-batch.

The results of semi-supervised classification experiments on MNIST and SVHN datasets are reported in Table \ref{table:semi}. On the MNIST dataset with 100 and 1000 labels, the performance of AAEs is significantly better than VAEs, on par with VAT \citep{vat} and CatGAN \citep{catgan}, but is outperformed by the Ladder networks \citep{ladder} and the ADGM \citep{adgm}. We also trained a supervised AAE model on all the available labels, and obtained the error rate of $0.85\%$. In comparison, a dropout supervised neural network with the same architecture achieves the error rate of $1.25\%$ on the full MNIST dataset, which highlights the regularization effect of the adversarial training. On the SVHN dataset with 1000 labels, the AAE almost matches the state-of-the-art classification performance achieved by the ADGM. 

It is also worth mentioning that all the AAE models are trained end-to-end, whereas the semi-supervised VAE models have to be trained one layer at a time \citep{semi-vae}. 

\begin{center}
\begin{table}[t]
\small
\centering 
\begin{tabular}{ l  || c | c | c || c}
  \hline
  & MNIST (100) & MNIST (1000) & MNIST (All) & SVHN (1000)\\
  \hline

  NN Baseline & $25.80$ & $8.73$ & $1.25$ & $47.50$\\
  \hline
  VAE (M1) + TSVM    & $11.82\;(\pm 0.25)$ & $4.24\;(\pm 0.07)$   & -                  & $55.33\;(\pm 0.11)$      \\
  VAE (M2)           & $11.97\;(\pm 1.71)$ & $3.60\;(\pm 0.56)$   & -                  & -                        \\
  VAE (M1 + M2)      & $3.33\;(\pm 0.14)$  & $2.40\;(\pm 0.02)$   & $0.96$             & $36.02\;(\pm 0.10)$      \\
  VAT                     & $2.33$              & $1.36$               & $0.64\;(\pm 0.04)$ & $24.63$                  \\
  CatGAN               & $1.91\;(\pm 0.1)$   & $1.73\;(\pm 0.18)$   & $0.91$             & -                        \\  
  Ladder Networks      & $1.06\;(\pm 0.37)$  & $0.84\;(\pm 0.08)$   & $0.57\;(\pm 0.02)$ & -                        \\
  ADGM                   & $0.96\;(\pm 0.02)$  & -                    & -                  & $16.61\;(\pm 0.24)$      \\
\hline 
  \textbf{Adversarial Autoencoders} & $1.90\;(\pm 0.10)$  & $1.60\;(\pm 0.08)$ & $0.85\;(\pm 0.02)$ & $17.70\;(\pm 0.30)$ \\
  \hline
\end{tabular}
\caption{\label{table:semi}Semi-supervised classification performance (error-rate) on MNIST and SVHN.}
\end{table}
\end{center}

\section{Unsupervised Clustering with Adversarial Autoencoders}
In the previous section, we showed that with a limited label information, the adversarial autoencoder is able to learn powerful semi-supervised representations. However, the question that has remained unanswered is whether it is possible to learn as ``powerful'' representations from unlabeled data without any supervision. In this section, we show that the adversarial autoencoder can disentangle discrete class variables from the continuous latent style variables in a purely unsupervised fashion. 

\begin{figure}[b]
\begin{center}
\centering 
\includegraphics[scale=.28]{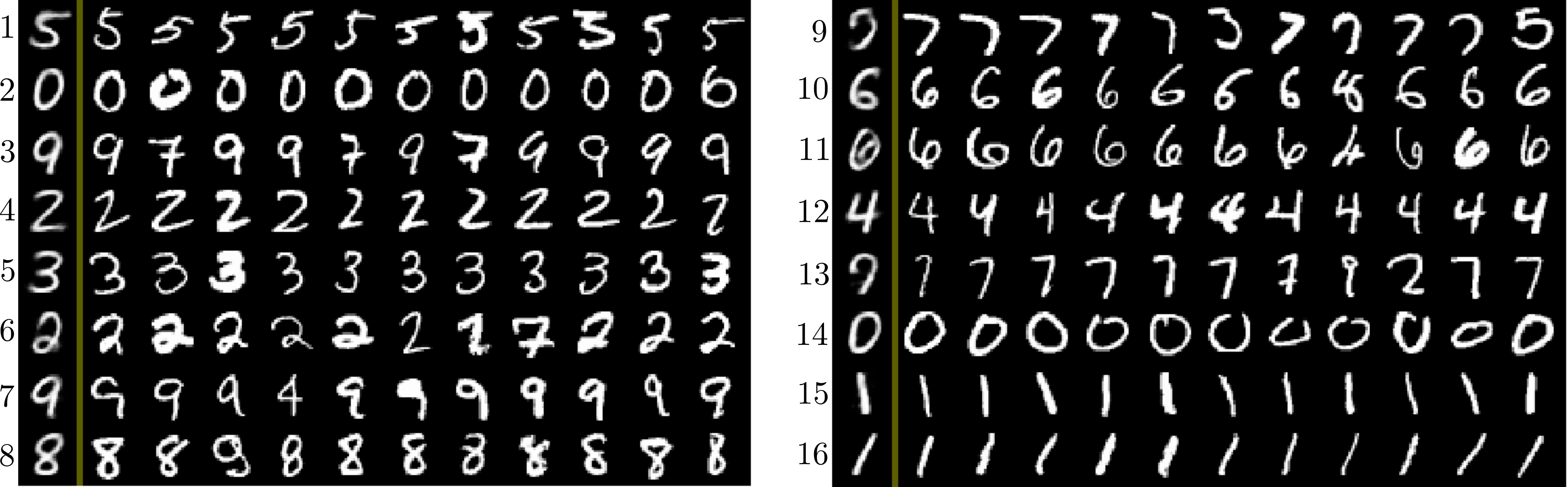}
\caption{\label{fig:cluster}Unsupervised clustering of MNIST using the AAE with 16 clusters. Each row corresponds to one cluster with the first image being the cluster head. (see text)}
\end{center}
\end{figure}

The architecture that we use is similar to Figure \ref{fig_semi_aae}, with the difference that we remove the semi-supervised classification stage and thus no longer train the network on any labeled mini-batch. Another difference is that the inference network $q(\mathbf{y}|\mathbf{x})$ predicts a one-hot vector whose dimension is the number of categories that we wish the data to be clustered into. Figure \ref{fig:cluster} illustrates the unsupervised clustering performance of the AAE on MNIST when the number of clusters is 16. Each row corresponds to one cluster. The first image in each row shows the cluster heads, which are digits generated by fixing the style variable to zero and setting the label variable to one of the 16 one-hot vectors. The rest of the images in each row are random test images that have been categorized into the corresponding category based on $q(\mathbf{y}|\mathbf{x})$. We can see that the AAE has picked up some discrete styles as the class labels. For example, the digit $1$s and $6$s that are tilted (cluster 16 and 11) are put in a separate cluster than the straight $1$s and $6$s (cluster 15 and 10), or the network has separated digit $2$s into two clusters (cluster 4, 6) depending on whether the digit is written with a loop. 

We performed an experiment to evaluate the unsupervised clustering performance of AAEs. We used the following evaluation protocol: Once the training is done, for each cluster $i$, we found the validation example $x_n$ that maximizes $q(y_i|x_n)$, and assigned the label of $x_n$ to all the points in the cluster $i$. We then computed the test error based on the assigned class labels to each cluster. As shown in Table \ref{table:cluster}, the AAE achieves the classification error rate of 9.55\% and 4.10\% with 16 and 30 total labels respectively. We observed that as the number of clusters grows, the classification rate improves.

\begin{center}
\begin{table}[t]
\small
\centering 
\begin{tabular}{ l  || c}
  \hline
  & MNIST (Unsupervised)\\
  \hline

  CatGAN \citep{catgan}(20 clusters) & $9.70$  \\  
  Adversarial Autoencoder (16 clusters) & $9.55\;(\pm 2.05)$ \\
  Adversarial Autoencoder (30 clusters) & $4.10\;(\pm 1.13)$ \\
  \hline
\end{tabular}
\caption{\label{table:cluster}Unsupervised clustering performance (error-rate) of the AAE on MNIST.}
\end{table}
\end{center}

\section{Dimensionality Reduction with Adversarial Autoencoders}\label{dim_reduce}

\begin{figure}[!b]
\begin{center}
\centering 
\includegraphics[scale=.4]{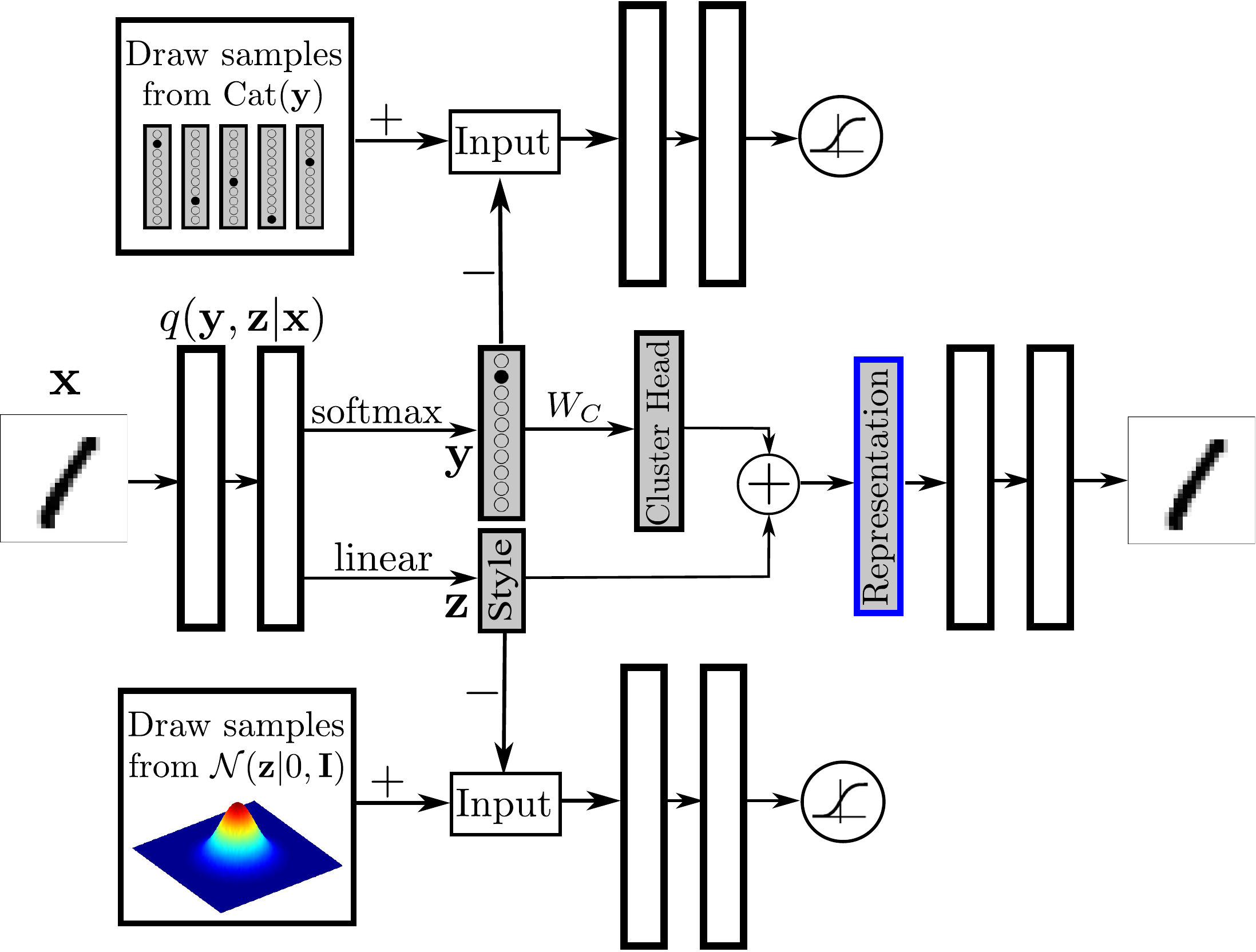}
\caption{\label{fig_2d}Dimensionality reduction with adversarial autoencoders: There are two separate adversarial networks that impose Categorical and Gaussian distribution on the latent representation. The final $n$ dimensional representation is constructed by first mapping the one-hot label representation to an $n$ dimensional cluster head representation and then adding the result to an $n$ dimensional style representation. The cluster heads are learned by SGD with an additional cost function that penalizes the Euclidean distance between of every two of them.}
\end{center}
\end{figure}

\begin{figure}[t]
\centering

\subfigure[2D representation with 1000 labels (4.20\% error)]{
\includegraphics[scale=.23]{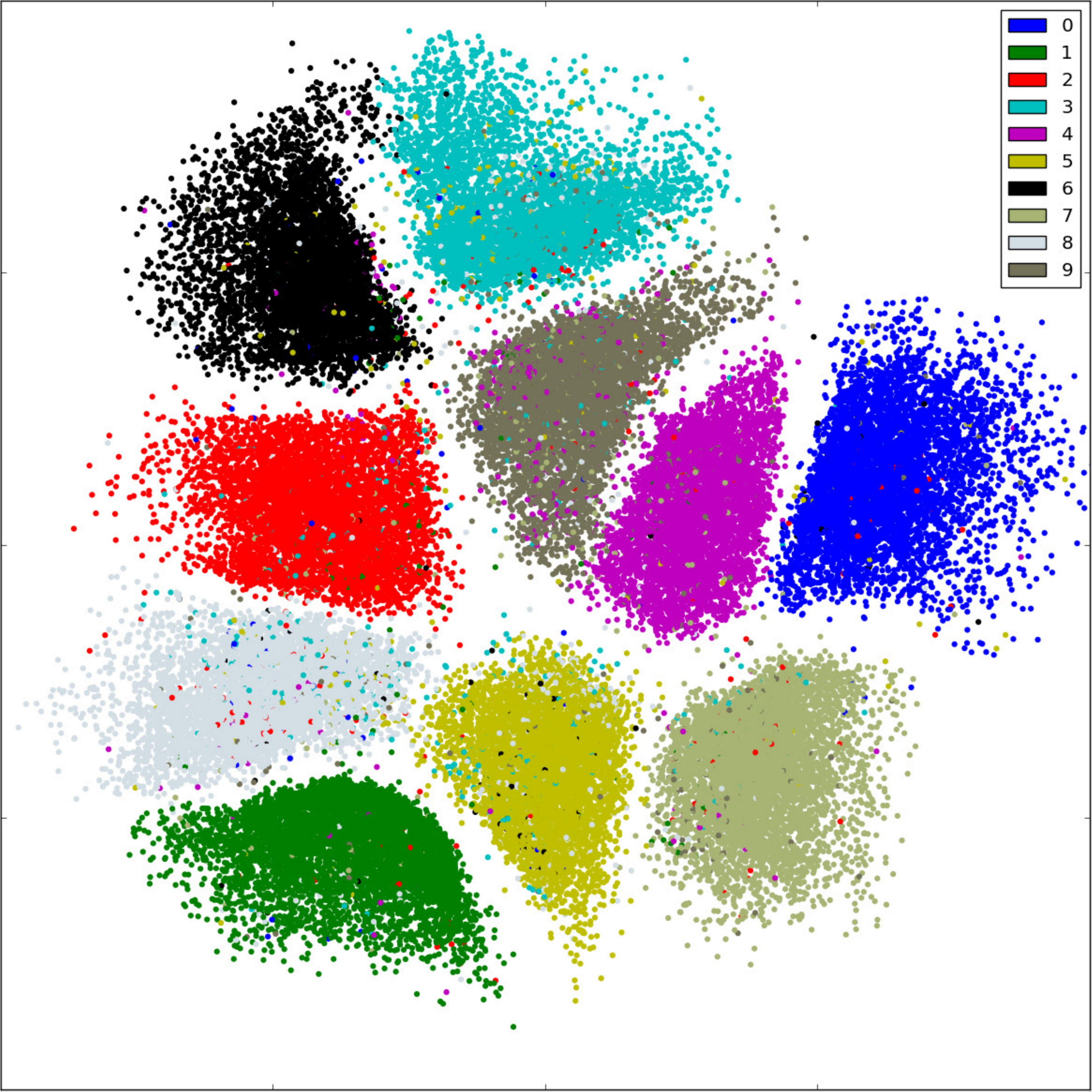}}
\subfigure[2D representation with 100 labels (6.08\% error)]{
\includegraphics[scale=.27]{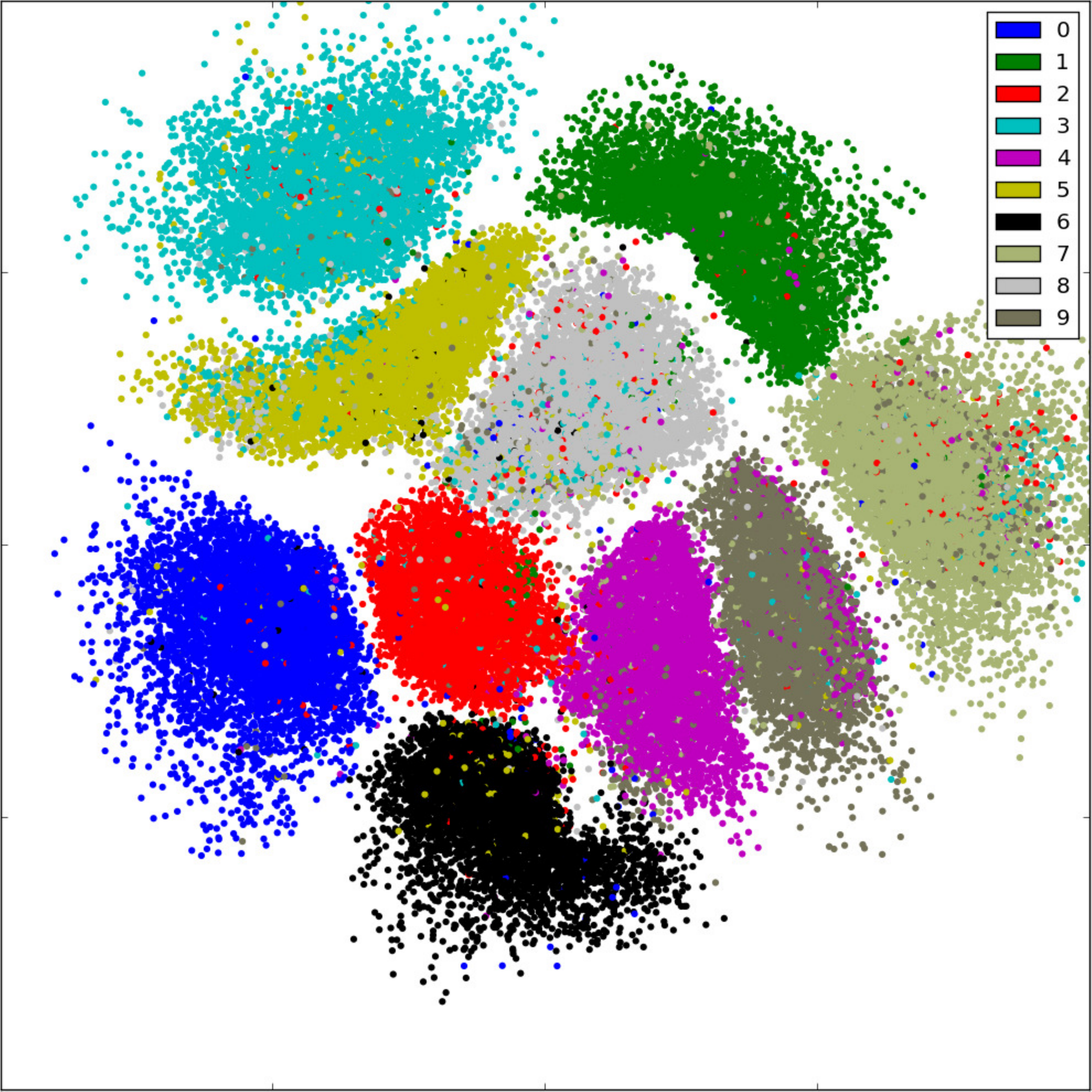}}
\subfigure[2D representation learnt in an unsupervised fashion with 20 clusters (13.95\% error)]{
\includegraphics[scale=.155]{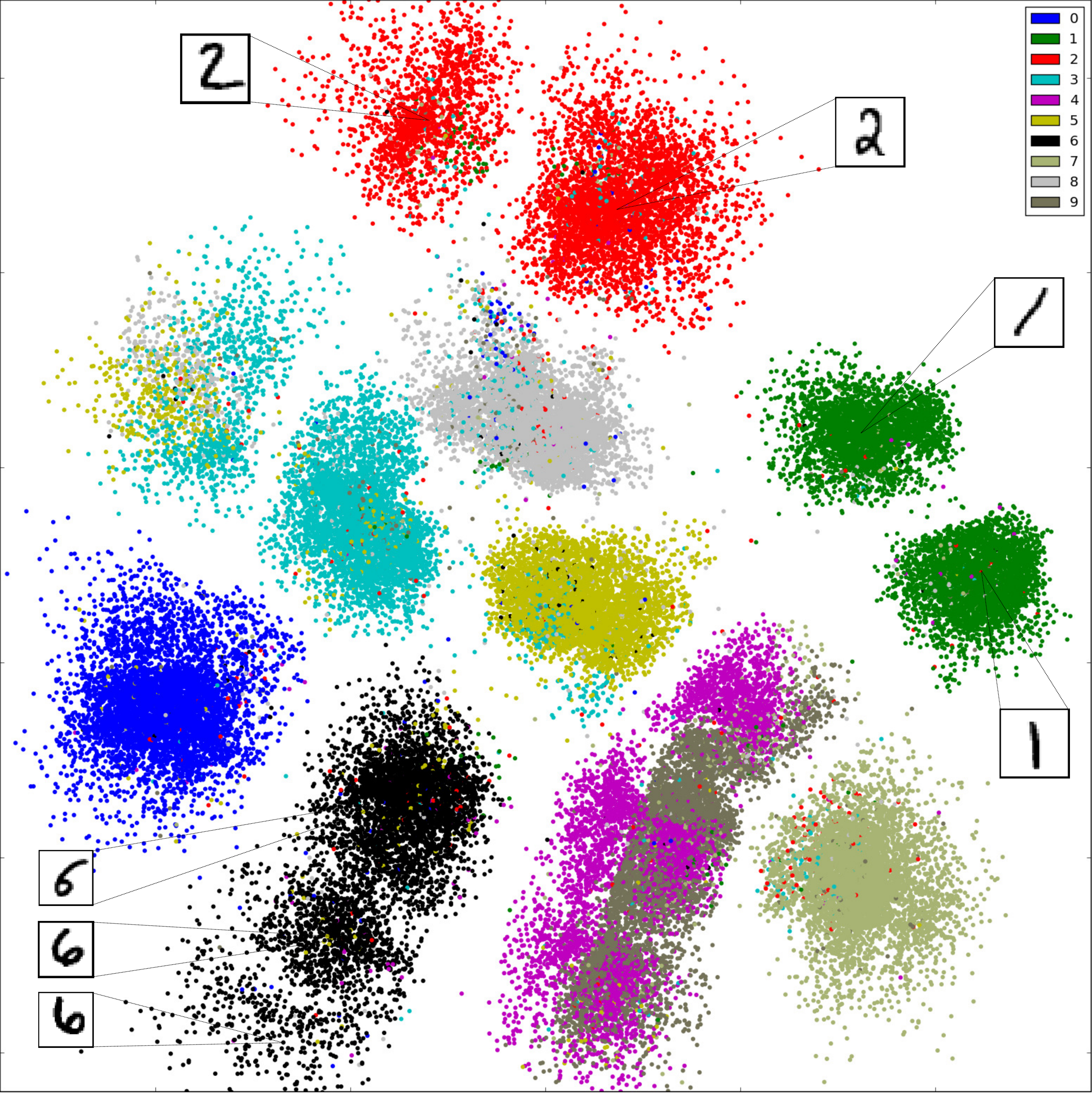}}
\subfigure[10D representation with 100 labels projected onto the 2D space (3.90\% error)]{
\includegraphics[scale=.235]{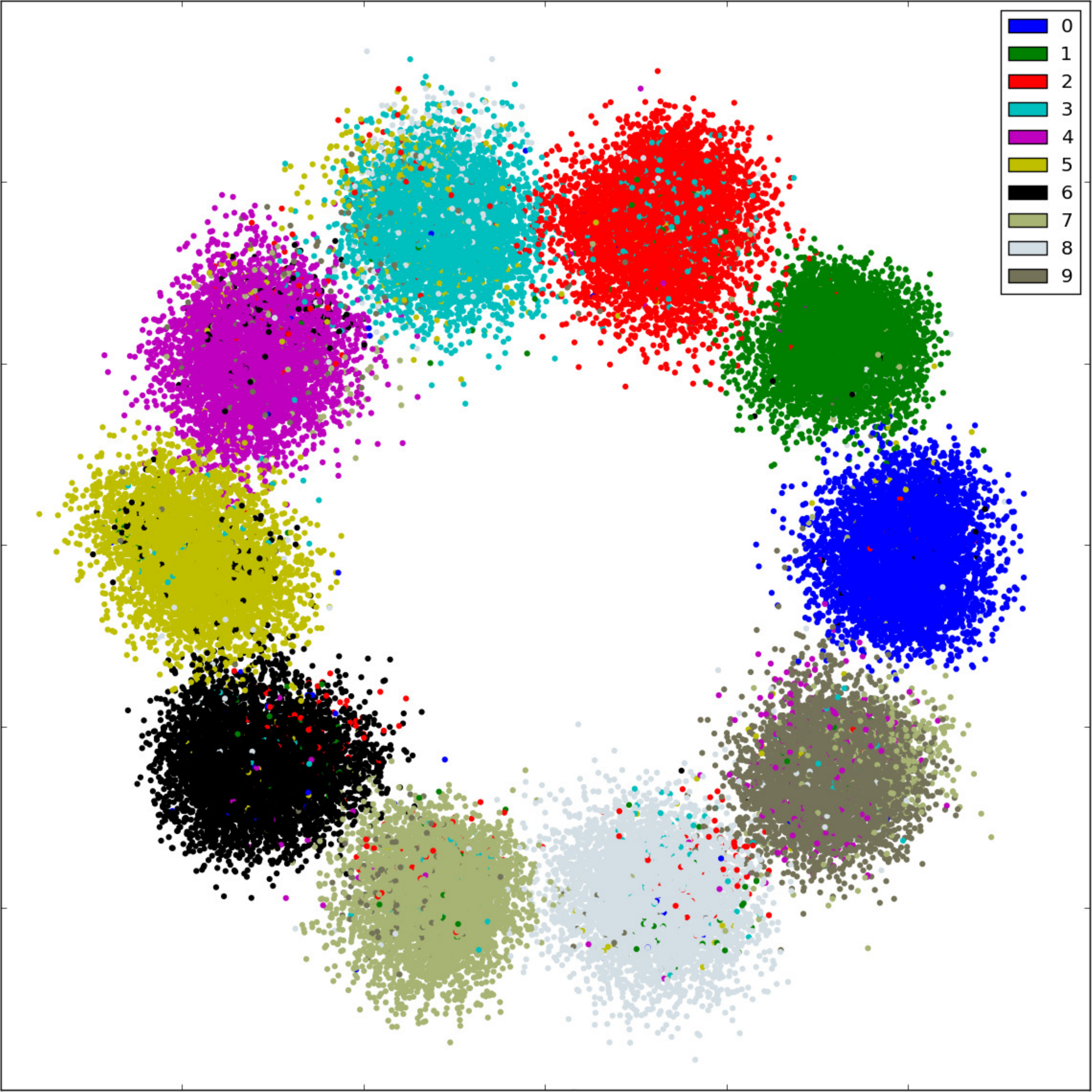}}
\caption{\label{fig:dim}Semi-Supervised and Unsupervised Dimensionality Reduction with AAE on MNIST.}
\end{figure}

Visualization of high dimensional data is a very important problem in many applications as it facilitates the understanding of the generative process of the data and allows us to extract useful information about the data. A popular approach of data visualization is learning a low dimensional embedding in which nearby points correspond to similar objects. Over the last decade, a large number of new non-parametric dimensionality reduction techniques such as t-SNE \citep{tsne} have been proposed. The main drawback of these methods is that they do not have a parametric encoder that can be used to find the embedding of the new data points. Different methods such as parametric t-SNE \citep{param_tsne} have been proposed to address this issue. Autoencoders are interesting alternatives as they provide the non-linear mapping required for such embeddings; but it is widely known that non-regularized autoencoders ``fracture'' the manifold into many different domains which result in very different codes for similar images \citep{geoff_dim_reduce}.

In this section, we present an adversarial autoencoder architecture for dimensionality reduction and data visualization purposes. We will show that in these autoencoders, the adversarial regularization attaches the hidden code of similar images to each other and thus prevents the manifold fracturing problem that is typically encountered in the embeddings learnt by the autoencoders. 

Suppose we have a dataset with $m$ class labels and we would like to reduce the dimensionality of the dataset to $n$, where $n$ is typically 2 or 3 for the visualization purposes. 
We alter the architecture of Figure \ref{fig_semi_aae} to Figure \ref{fig_2d} in which the final representation is achieved by adding the $n$ dimensional distributed representation of the cluster head with the $n$ dimensional style representation. The cluster head representation is obtained by multiplying the $m$ dimensional one-hot class label vector by an $m \times n$ matrix $W_C$, where the rows of $W_C$ represent the $m$ cluster head representations that are learned with SGD. We introduce an additional cost function that penalizes the Euclidean distance between every two cluster heads. Specifically, if the Euclidean distance is larger than a threshold $\eta$, the cost function is zero, and if it is smaller than $\eta$, the cost function linearly penalizes the distance.

Figure \ref{fig:dim} (a, b) show the results of the semi-supervised dimensionality reduction in $n=2$ dimensions on the MNIST dataset ($m=10$) with 1000 and 100 labels. We can see that the network can achieve a clean separation of the digit clusters and obtain the semi-supervised classification error of 4.20\% and 6.08\% respectively. Note that because of the 2D constraint, the classification error is not as good as the high-dimensional cases; and that the style distribution of each cluster is not quite Gaussian.

Figure \ref{fig:dim}c shows the result of unsupervised dimensionality reduction in $n=2$ dimensions where the number of clusters have chosen to be $m=20$. We can see that the network can achieve a rather clean separation of the digit clusters and sub-clusters. For example, the network has assigned two different clusters to digit 1 (green clusters) depending on whether the digit is straight or tilted. The network is also clustering digit 6 into three clusters (black clusters) depending on how much tilted the digit is. Also the network has assigned two separate clusters for digit 2 (red clusters), depending on whether the digit is written with a loop.

This AAE architecture (Figure \ref{fig_2d}) can also be used to embed images into larger dimensionalities ($n>2$). For example, Figure \ref{fig:dim}d shows the result of semi-supervised dimensionality reduction in $n=10$ dimensions with 100 labels. In this case, we fixed $W_C$ matrix to $W_C = 10\mathbf{I}$ and thus the cluster heads are the corners of a $10$ dimensional simplex. The style representation is learnt to be a 10D Gaussian distribution with the standard deviation of 1 and is directly added to the cluster head to construct the final representation. Once the network is trained, in order to visualize the 10D learnt representation, we use a linear transformation to map the 10D representation to a 2D space such that the cluster heads are mapped to the points that are uniformly placed on a 2D circle. We can verify from this figure that in this high-dimensional case, the style representation has indeed learnt to have a Gaussian distribution. With 100 total labels, this model achieves the classification error-rate of 3.90\% which is worse than the classification error-rate of 1.90\% that is achieved by the AAE architecture with the concatenated style and label representation (Figure \ref{fig_semi_aae}).

\section{Conclusion}

In this paper, we proposed to use the GAN framework as a variational inference algorithm for both discrete and continuous latent variables in probabilistic autoencoders. Our method called the adversarial autoencoder (AAE), is a generative autoencoder that achieves competitive test likelihoods on real-valued MNIST and Toronto Face datasets. We discussed how this method can be extended to semi-supervised scenarios and showed that it achieves competitive semi-supervised classification performance on MNIST and SVHN datasets. Finally, we demonstrated the applications of adversarial autoencoders in disentangling the style and content of images, unsupervised clustering, dimensionality reduction and data visualization.

\subsubsection*{Acknowledgments}
We would like to thank Ilya Sutskever, Oriol Vinyals, Jon Gauthier, Sam Bowman and other members of the Google Brain team for helpful discussions. We thank the developers of TensorFlow \citep{tensorflow2015-whitepaper}, which we used for all of our experiments. We also thank NVIDIA for GPU donations. 

\bibliographystyle{unsrtnat}
\bibliography{nips_2016}

\begin{appendices}

\section{Experiment Details}\label{exp:details}

\subsection{Likelihood Experiments}
The encoder, decoder and discriminator each have two layers of 1000 hidden units with ReLU activation function. The activation of the last layer of $q(\mathbf{z}|\mathbf{x})$ is linear. 
The weights are initialized with a Gaussian distribution with the standard deviation of 0.01. 
The mini-batch size is 100.
The autoencoder is trained with a Euclidean cost function for reconstruction.
On the MNIST dataset we use the sigmoid activation function in the last layer of the autoencoder and on the TFD dataset we use the linear activation function.
The dimensionality of the hidden code $\mathbf{z}$ is $8$ and $15$ and the standard deviation of the Gaussian prior is $5$ and $10$ for MNIST and TFD, respectively. On the Toronto Face dataset, data points are subtracted by the mean and divided by the standard deviation along each input dimension across the whole training set to normalize the contrast. However, after obtaining the samples, we rescaled the images (by inverting the pre-processing stage) to have pixel intensities between 0 and 1 so that we can have a fair likelihood comparison with other methods.

In the deterministic case of $q(\mathbf{z}|\mathbf{x})$, the dimensionality of the hidden code should be consistent with the intrinsic dimensionality of the data, since the only source of stochasticity in $q(\mathbf{z})$ is the data distribution. For example, in the case of MNIST, the dimensionality of the hidden code can be between 5 to 8, and for TFD and SVHN, it can be between 10 to 20. For training AAEs with higher dimensionalities in the code space (e.g., 1000), the probabilistic $q(\mathbf{z}|\mathbf{x})$ along with the re-parametrization trick can be used.

\subsection{Semi-Supervised Experiments}
\subsubsection{MNIST}\label{mnist-semi}
The encoder, decoder and discriminator each have two layers of 1000 hidden units with ReLU activation function. The last layer of the autoencoder can have a linear or sigmoid activation (sigmoid is better for sample visualization). The cost function is half the Euclidean error. The last layer of $q(\mathbf{y}|\mathbf{x})$ and $q(\mathbf{z}|\mathbf{x})$ has the softmax and linear activation function, respectively. The $q(\mathbf{y}|\mathbf{x})$ and $q(\mathbf{z}|\mathbf{x})$ share the first two 1000-unit layers of the encoder. The dimensionality of both the style and label representation is 10. On the style representation, we impose a Gaussian distribution with the standard deviation of 1. On the label representation, we impose a Categorical distribution. The semi-supervised cost is a cross-entropy cost function at the output of $q(\mathbf{y}|\mathbf{x})$. We use gradient descent with momentum for optimizing all the cost functions. The momentum value for the autoencoder reconstruction cost and the semi-supervised cost is fixed to 0.9. The momentum value for the generator and discriminator of both of the adversarial networks is fixed to 0.1. For the reconstruction cost, we use the initial learning rate of 0.01, after 50 epochs reduce it to 0.001 and after 1000 epochs reduce it to 0.0001. For the semi-supervised cost, we use the initial learning rate of 0.1, after 50 epochs reduce it to 0.01 and after 1000 epochs reduce it to 0.001. For both the discriminative and generative costs of the adversarial networks, we use the initial learning rate of 0.1, after 50 epochs reduce it to 0.01 and after 1000 epochs reduce it to 0.001. We train the network for 5000 epochs. We add a Gaussian noise with standard deviation of 0.3 only to the input layer and only at the training time. No dropout, $\ell_2$ weight decay or other Gaussian noise regularization were used in any other layer. The labeled examples were chosen at random, but we made sure they are distributed evenly across the classes. In the case of MNIST with 100 labels, the test error after the first epochs is 16.50\%, after 50 epochs is 3.40\%, after 500 epochs is 2.21\% and after 5000 epochs is 1.80\%. Batch-normalization \citep{batch} did not help in the case of the MNIST dataset.

\subsubsection{SVHN}
The SVHN dataset has about 530K training points and 26K test points. Data points are subtracted by the mean and divided by the standard deviation along each input dimension across the whole training set to normalize the contrast. The dimensionality of the label representation is 10 and for the style representation we use 20 dimensions. We use gradient descent with momentum for optimizing all the cost functions. The momentum value for the autoencoder reconstruction cost and the semi-supervised cost is fixed to 0.9. The momentum value for the generator and discriminator of both of the adversarial networks is fixed to 0.1. For the reconstruction cost, we use the initial learning rate of 0.0001 and after 250 epochs reduce it to 0.00001. For the semi-supervised cost, we use the initial learning rate of 0.1 and after 250 epochs reduce it to 0.01. For both the discriminative and generative costs of the adversarial networks, we use the initial learning rate of 0.01 and after 250 epochs reduce it to 0.001. We train the network for 1000 epochs. We use dropout at the input layer with the dropout rate of 20\%. No other dropout, $\ell_2$ weight decay or Gaussian noise regularization were used in any other layer. The labeled examples were chosen at random, but we made sure they are distributed evenly across the classes. In the case of SVHN with 1000 labels, the test error after the first epochs is 49.34\%, after 50 epochs is 25.86\%, after 500 epochs is 18.15\% and after 1000 epochs is 17.66\%. Batch-normalization were used in all the autoencoder layers including the softmax layer of $q(\mathbf{y}|\mathbf{x})$, the linear layer of $q(\mathbf{z}|\mathbf{x})$ as well as the linear output layer of the autoencoder. We found batch-normalization \citep{batch} to be crucial in training the AAE network on the SVHN dataset.

\subsection{Unsupervised Clustering Experiments}
The encoder, decoder and discriminator each have two layers of 3000 hidden units with ReLU activation function. The last layer of the autoencoder has a sigmoid activation function. The cost function is half the Euclidean error. The dimensionality of the style and label representation is 5 and 30 (number of clusters), respectively. On the style representation, we impose a Gaussian distribution with the standard deviation of 1. On the label representation, we impose a Categorical distribution. We use gradient descent with momentum for optimizing all the cost functions. The momentum value for the autoencoder reconstruction cost is fixed to 0.9. The momentum value for the generator and discriminator of both of the adversarial networks is fixed to 0.1. For the reconstruction cost, we use the initial learning rate of 0.01 and after 50 epochs reduce it to 0.001. For both the discriminative and generative costs of the adversarial networks, we use the initial learning rate of 0.1 and after 50 epochs reduce it to 0.01. We train the network for 1500 epochs. We use dropout at the input layer with the dropout rate of 20\%. No other dropout, $\ell_2$ weight decay or Gaussian noise regularization were used in any other layer. Batch-normalization was used only in the encoder layers of the autoencoder including the last layer of $q(\mathbf{y}|\mathbf{x})$ and $q(\mathbf{z}|\mathbf{x})$. We found batch-normalization \citep{batch} to be crucial in training the AAE networks for unsupervised clustering.

\end{appendices}

\end{document}